\documentclass[10pt,twocolumn,letterpaper]{article}

\usepackage[pagenumbers]{iccv} % To force page numbers, e.g. for an arXiv version

\definecolor{iccvblue}{rgb}{0.21,0.49,0.74}
\usepackage[pagebackref,breaklinks,colorlinks,allcolors=iccvblue]{hyperref}
\usepackage{algorithm}
\usepackage{algorithmic}
\usepackage{multicol}
\usepackage{multirow}
\usepackage{booktabs}
\usepackage{setspace}
\usepackage{amsmath}
\usepackage{amsfonts}
\usepackage{CJK}
\usepackage{indentfirst}
\usepackage{bm}
\usepackage{graphicx}
\usepackage[section]{placeins}
\usepackage{makecell}
\makeatletter
\let\MYcaption\@makecaption
\makeatother

\usepackage[font=footnotesize]{subcaption}
\usepackage{multirow}
\usepackage{amssymb}
\usepackage{color}
\usepackage{amsthm}
\usepackage[font=footnotesize]{subcaption}
\usepackage{xr-hyper}
\usepackage{hyperref}
\theoremstyle{plain}
\newtheorem{theorem}{Theorem}[section]

\theoremstyle{definition}

\theoremstyle{remark}

\def\paperID{268} % *** Enter the Paper ID here
\def\confName{ICCV}
\def\confYear{2025}

\title{BaryIR: Learning Multi-Source Unified Representation in Continuous Barycenter Space for Generalizable All-in-One Image Restoration}

\author{Xiaole Tang~~~~~~Xiaoyi He~~~~~~Xiang Gu~~~~~~Jian Sun\\
	Xi'an Jiaotong University, Xi'an, China\\
	{\tt\small \{tangxl,hexiaoyi\}@stu.xjtu.edu.cn, \{xianggu,jiansun\}@xjtu.edu.cn}
	}
\begin{document}
\maketitle
\begin{abstract}
	%	The map is derived and parameterized by a lightweight neural network based on the dual formulation of the optimal transport (OT) barycenter problem,  which seeks the ``average'' distribution aggregating the multi-source latent ones.
Despite remarkable advances made in all-in-one image restoration (AIR) for handling different types of degradations simultaneously, existing methods remain vulnerable to out-of-distribution degradations and images, limiting their real-world applicability. In this paper, we propose a multi-source representation learning framework BaryIR, which decomposes the latent space of multi-source degraded images into a continuous barycenter space for unified feature encoding and source-specific subspaces for specific semantic encoding.  Specifically, we seek the multi-source unified representation by introducing a multi-source latent optimal transport barycenter problem, in which a continuous barycenter map is learned to transport the latent representations to the barycenter space. The transport cost is designed such that the representations from source-specific subspaces are contrasted with each other while maintaining orthogonality to those from the barycenter space. This enables BaryIR to learn compact representations with unified degradation-agnostic information from the barycenter space, as well as degradation-specific semantics from source-specific subspaces, capturing the inherent geometry of multi-source data manifold for generalizable AIR. Extensive experiments demonstrate that BaryIR achieves competitive performance compared to state-of-the-art all-in-one methods. Particularly, BaryIR exhibits superior generalization ability to real-world data and unseen degradations. The code will be publicly available at \url{https://github.com/xl-tang3/BaryIR}. \vspace{-0.4cm}

\end{abstract}
\section{Introduction}
\begin{figure}[!t]
	\centering
	\includegraphics[width=1\linewidth]{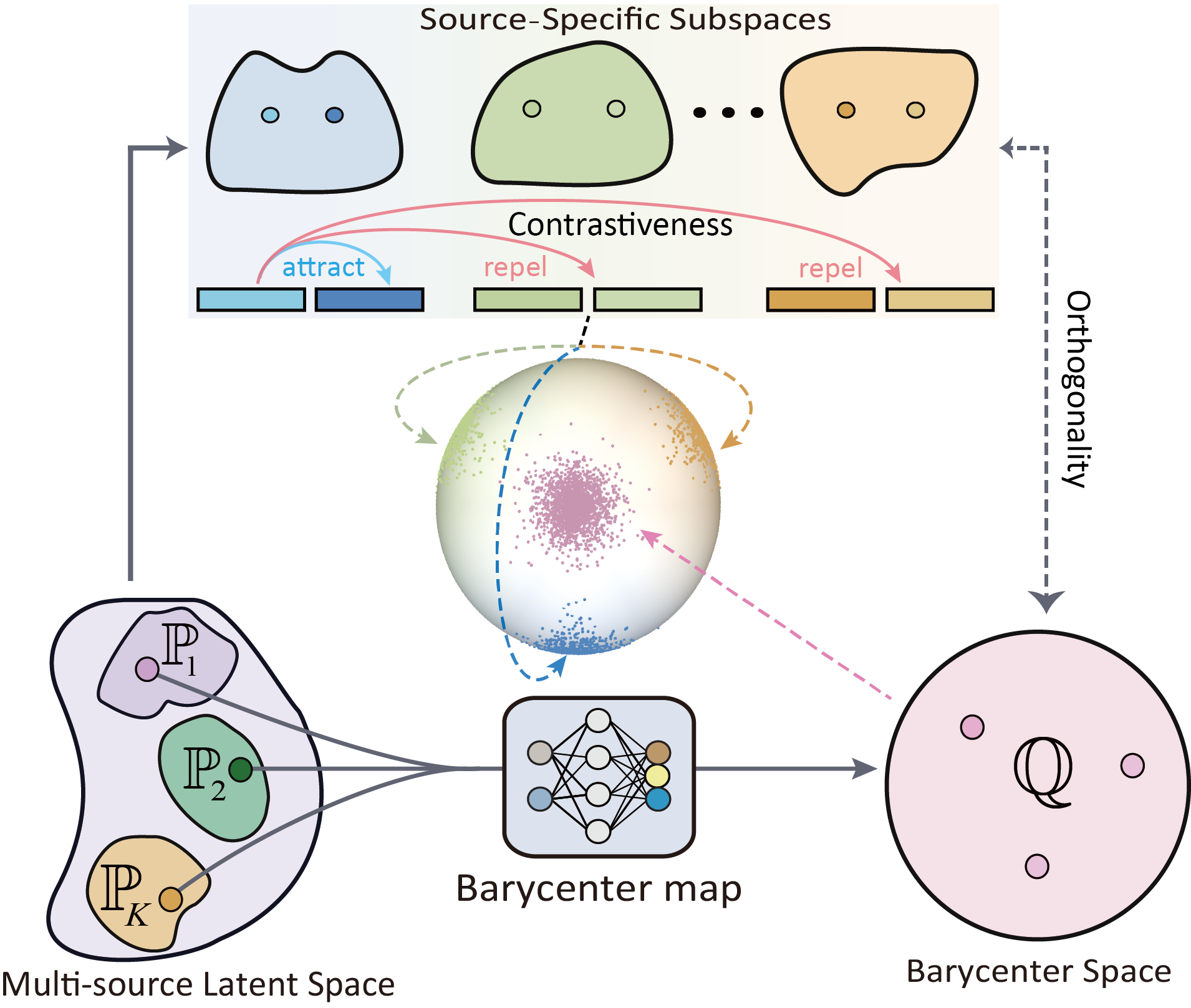}
	\vspace{-0.6cm}
	\caption{BaryIR decomposes the latent space of multi-source degraded images into a continuous barycenter space and source-specific subspaces. The source-specific representations are contrasted with each other while remaining orthogonal to the barycenter ones. The barycenter space seeks to encode degradation-agnostic features by aggregating the multiple source domains, which enriches the overall geometry of the data manifold. }
	\label{demo}
	\vspace{-0.6cm}
\end{figure}
Image restoration plays a fundamental role in low-level vision, aiming to recover the high-quality images given the degraded counterparts. Recent advances of deep neural networks (NNs) \cite{he2016deep, vaswani2017attention, dosovitskiy2021an, liu2021swin} have triggered remarkable successes in image restoration, in which most works \cite{lai2017deep, lai2018fast, zhang2018ffdnet, Zamir2021MPRNet, Zamir2021Restormer, liang2021swinir, chen2022simple, Tang_2023_CVPR, luo2023image, zhou2023fourmer} develop task-specific restoration networks to handle single known degradations (\textit{e.g.,} noise, blur, rain, haze, low light). However, this specificity hinders their applicability in real-world scenarios such as autonomous navigation \cite{levinson2011towards, prakash2021multi} and surveillance systems \cite{liang2018deep}, where varied and unexpected degradations frequently occur. Consequently, there has been emerging interest in addressing multiple forms of degradations within a single model, known as all-in-one image restoration (AIR) \cite{jiang2024survey}.

In response to the AIR problem, most existing works  \cite{li2022all, valanarasu2022transweather, zhang2023ingredient, potlapalli2023promptir, luocontrolling, tang2024residualconditioned, cui2025adair} leverage degradation-specific information to guide the unified restoration networks by encoding extra degradation-specific signals, \textit{e.g.,} learnable prompts \cite{potlapalli2023promptir, luocontrolling, valanarasu2022transweather, liu24k}, residual embeddings \cite{tang2024residualconditioned, tang2025degradation}, and frequency bands \cite{cui2025adair}. Some other works \cite{conde2024instructir, zamfir2024efficient, ai2024lora} adopt mixture-of-experts or adaptation modules to process images with different degradations, thereby leading to degradation-aware restoration. 
Although these methods integrate degradation-specific dynamics into the restoration network, they often struggle to capture the degradation-agnostic features of the multi-source degraded images, which are crucial for learning general commonality beyond the training samples. Consequently, they remain vulnerable to out-of-distribution (OOD) degraded images, limiting their generalization performance in real-world scenarios.

In this paper, we address the AIR problem by introducing a multi-source representation learning framework BaryIR, which decomposes the latent space of multi-source degraded images into a continuous barycenter space for unified encoding, and source-specific subspaces that encode degradation semantics as the restoration guidance (Fig. \ref{demo}). Specifically, we learn a continuous optimal transport (OT) barycenter map that transports representations to the continuous barycenter space, where the multi-source representations are aligned. The map is derived and parameterized by a lightweight NN based on the dual reformulation of the OT barycenter problem, which seeks the optimal ``average'' distribution that aggregates the multi-source latent distributions while mitigating the training imbalance among different sources. The barycenter problem is formulated through a multi-source latent optimal transport (MLOT) objective, which exploits the source-level contrastiveness among source-specific subspaces, while imposing orthogonality between the barycenter and these subspaces (see Fig. \ref{demo}).  This approach enables compact decomposition of representations into the barycenter and source-specific ones, resulting in generalizable AIR representations that capture the inherent geometry of multi-source degraded images.

In summary, our contributions are as follows:
\begin{itemize}
	\item We present BaryIR, a novel framework that seeks continuous barycenters for multi-source unified representation. By decomposing the latent space into continuous barycenter space and source-specific subspaces, BaryIR captures the inherent geometry of multi-source data for generalizable all-in-one image restoration.
	
	\item With the dual formulation of the MLOT barycenter problem, we learn an NN-based barycenter map that transports representations to the barycenter space for unified encoding, which alleviates the training imbalance among different degradations. Moreover, we theoretically establish the error bounds for the barycenter map, providing guarantees on its approximation quality.
	
	\item Extensive experiments on both synthetic and real-world data show that BaryIR achieves state-of-the-art performance in all-in-one and task-specific image restoration. Notably, BaryIR exhibits superior generalization ability to unseen degradations and real-world data.
\end{itemize}

\section{Related Work}
\textbf{All-in-One Image Restoration.} Pioneer AIR methods typically utilize informative degradation embeddings \cite{li2022all, zamfir2024efficient, potlapalli2023promptir, luocontrolling, tang2024residualconditioned, chen2024learning, cui2025adair} to guide the restoration. For instance, AirNet \cite{li2022all} trains an extra encoder using contrastive learning to extract degradation embeddings from degraded images.  PromptIR \cite{potlapalli2023promptir} and DA-CLIP \cite{luocontrolling} employ learnable visual prompts to encode the information of degradation type. Another line of works, \textit{e.g.,} InstructIR \cite{conde2024instructir}, DaAIR \cite{zamfir2024efficient}, Histoformer \cite{sun2024restoring},  route samples with different degradation patterns to specific experts or architectures for dynamic restoration. However, these approaches are vulnerable to OOD degradations (\textit{e.g.,} unseen degradation patterns and levels) and are hardly generalizable due to the difficulty in capturing general and intrinsic commonality among the source domains. In contrast, BaryIR seeks to decompose the latent space into a barycenter space and source-specific subspaces, allowing us to explicitly learn compact unified and source-specific representations that capture the comprehensive geometry of the multi-source degraded images for generalizable AIR.

\textbf{Unified Representation Learning.} Learning unified representations is a fundamental aspect of multimodal or multi-view learning. The majority of existing works aim to align diverse sources/modalities (\textit{e.g.,} text and images) within a shared latent space \cite{radford2021learning, sarkar2024xkd, andonian2022robust, xue2023ulip} or train a source-agnostic encoder to extract information across heterogeneous sources \cite{chen2020uniter, wang2022vlmixer}. The other line of works explores how to express the shared content from different domains with explicit unified representations, \textit{e.g.,} codebooks \cite{lu2023unifiedio, liu-etal-2022-cross, ao2022speecht5} or prototypes \cite{yang2023prototypical, duan2022multi}. For example, Duan et al. \cite{duan2022multi} employ discrete OT to map the features extracted from different modalities to the prototypes. Despite their successes, these methods typically project feature vectors into a unified discrete space, which inherently limits their ability to capture the high-dimensional, fine-grained structures of the data manifold. In this paper, we explore how to learn unified representation in the continuous barycenter space.
\vspace{-0.2cm}
\section{Preliminaries}
\vspace{-0.1cm}
\textbf{Notation.} In this paper, we denote $\bar K=\{1,2,\dots,K\}$ for $K\in \mathbb N$. Given elements \( e_1, e_2, \ldots \) indexed by natural numbers, we denote the tuple \((e_1, e_2, \ldots, e_K)\) as \( e_{1:K} \). $\mathcal X\subset \mathbb R^d, \mathcal Y\subset \mathbb R^{d'},\mathcal X_k\subset\mathbb R^{d_k}$  are compact subsets of Euclidean space. $\mathcal C(\mathcal X)$ is the space of continuous functions on $\mathcal X$. The set of distributions on $\mathcal X$ is denoted by $\mathcal P(\mathcal X)$. For $\mathbb P \in \mathcal P(\mathcal X)$,  $\mathbb Q \in \mathcal P(\mathcal Y)$, the set of \textit{transport plans} is denoted as $\Pi(\mathbb P, \mathbb Q)$, \textit{i.e.,} probability distributions on $\mathcal X \times \mathcal Y$ with first and second marginals $\mathbb P$ and $\mathbb Q$. The pushforward of distribution $\mathbb P$ under some measurable map $T$ is denoted by $T_\#\mathbb P$. The Operator $\langle\cdot,\cdot\rangle$ denotes the cosine similarity that involves the normalization of features (on the unit sphere).
\begin{figure*}[!t]
	\centering
	\includegraphics[width=1\linewidth]{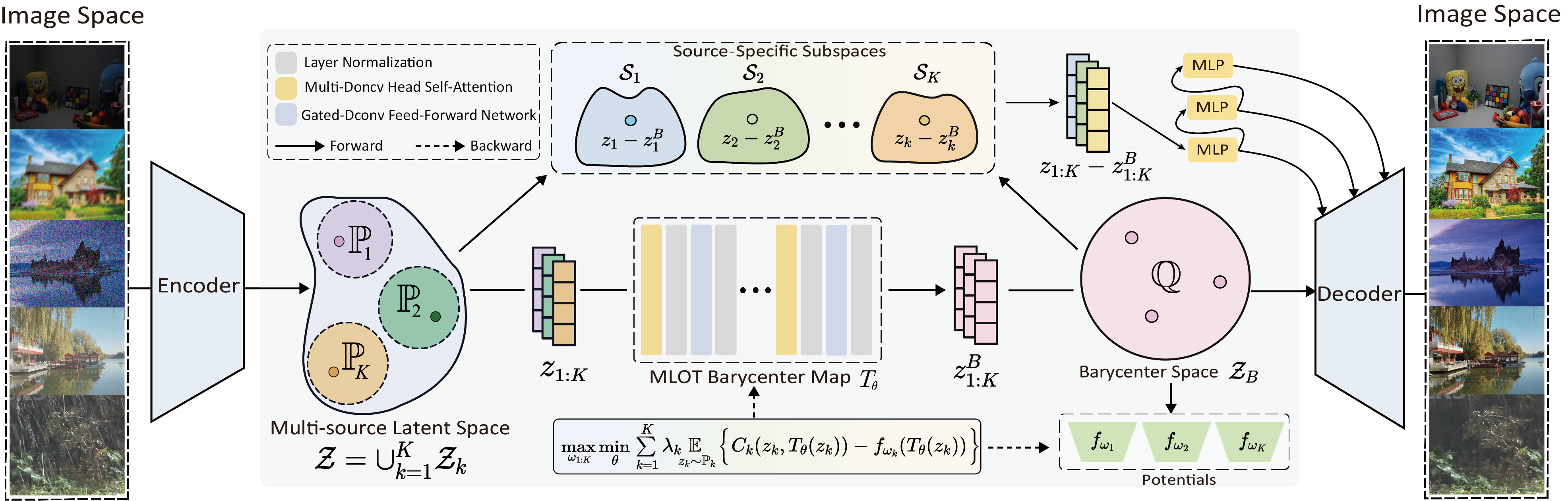}
	\caption{\textbf{Overview of the proposed BaryIR framework.} Based on the MLOT barycenter objective, we train the MLOT barycenter map that transports the latent representation to the \textit{barycenter space}. Correspondingly, we can establish the \textit{source-specific subspaces}  with elements being differences between the sources and barycenters. By aggregating representations from both spaces, BaryIR can capture degradation-agnostic/specific semantics for all-in-one image restoration. The encoder and decoder adopts the Restormer \cite{Zamir2021Restormer} architecture.}
	\label{pipeline}
	\vspace{-0.2cm}
\end{figure*}
\subsection{Optimal Transport}
Given two distributions $\mathbb P\in \mathcal P(\mathcal Y)$ and $\mathbb Q\in\mathcal P(\mathcal X)$ with a transport cost function $c: \mathcal X \times \mathcal Y \rightarrow \mathbb{R}_+$, the Kantorovich formulation \cite{kantorovich1942translocation} of the OT problem is defined as:
\begin{align}
	\text{OT}_c(\mathbb{P},\mathbb{Q})\triangleq\inf_{\pi\in\Pi(\mathbb P,\mathbb Q)}\int_{\mathcal X\times \mathcal Y}c(x,y)d\pi(x,y),
	\label{Kon}
\end{align}
where $\pi\in\Pi(\mathbb P,\mathbb Q)$ is a transport plan. The choice of $c(x,y)=\|x-y\|$ yields the Earth Mover's Distance.  The plan $\pi^*$ attaining the infimum is the \textit{optimal transport plan}. The problem (\ref{Kon}) admits the following dual form \cite{villani2009optimal}:
\begin{align}
	\text{OT}_c(\mathbb{P},\mathbb{Q})=\sup_{f}\int_{\mathcal X} f^{c}(x)d\mathbb P(x)+\int_{\mathcal Y} f(y)d\mathbb Q(y),
	\label{dual}
\end{align}
where $\displaystyle f^{c}(x)=\inf_{y\in \mathcal Y}\left[c(x,y)-f(y)\right]$ is the $c$-transform of the potential function $f\in\mathcal C(\mathcal Y)$.
\subsection{Classic Optimal Transport Barycenter}
Given distributions $\mathbb P_k\in\mathcal P(\mathcal X_k)$ for $k\in\bar K$ and transport costs $c_k:\mathcal X_k\times \mathcal Y\rightarrow \mathbb R_+$. For weights $\lambda_k>0$ with $\sum_{k=1}^K\lambda_k=1$, the classic OT barycenter problem seeks the distribution $\mathbb Q$ that attains the minimum of the weighted sum of OT problems with fixed first marginals $\mathbb P_{1:K}$:
\begin{align}
	\inf_{\mathbb Q\in\mathcal P(\mathcal Y)}\sum_{k=1}^{K}\lambda_k\text{OT}_{c_k}(\mathbb P_k, \mathbb Q). \label{classic-bary}
\end{align}
In practice, given $N_k$ empirical samples $x_{1:N_k}^k\sim\mathbb P_k$ in a multi-source space $\mathcal X=\cup_{k=1}^K\mathcal X_k$, the distributions $\mathbb P_{k}$ for $\mathcal X_k$ can be assessed using these empirical samples. Based on the OT barycenter problem (\ref{classic-bary}), we can establish a map $ T:\mathcal X\rightarrow \mathcal Y$, which allows sampling points $T(x_k)$ from the approximate barycenter space with $x_k \sim \mathbb{P}_k$ as inputs. The setup leads to a continuous barycenter problem. Different from prior works \cite{li2020continuous, korotin2021continuous, chi2023variational} that model individual maps for each source and test on simple domains, we seek the unified representation of high-dimensional multi-source data by learning an NN-based unified barycenter map.
\section{Method}
We present BaryIR, which tackles AIR by decomposing the multi-source latent space of degraded images into a continuous barycenter space that encodes unified features with OT barycenters and source-specific subspaces that provide degradation-specific semantics. This decomposition allows BaryIR to capture degradation-agnostic and degradation-specific features separately, better representing the underlying geometry of the multi-source data.

\textbf{Overview.} We seek continuous latent barycenters for unified representation of multi-source degraded images with the MLOT objective (\S\ref{4.1}). Based on the dual reformulation of the MLOT barycenter problem, we optimize to learn an NN-based barycenter map to transport multi-source latent representations to the barycenter space for unified encoding and exploit the source-specific subspaces for degradation-specific semantic encoding (in \S\ref{4.2} and \S\ref{4.3}). The representations from both spaces are combined as aggregated features of multi-source degraded images, which are then decoded into clear images (Fig. \ref{pipeline}).
%\S\ref{4.1} introduces the latent barycenter problem with multi-source latent OT (MLOT) objective for multi-source unified representation learning. In \S\ref{4.2}, we derive and parameterize the barycenter map based on the dual formulation of MLOT barycenter problem and establish its error bounds. In \S\ref{4.3}, we present the practical implementation with NNs and the computational algorithm for the MLOT objective. 

%a novel transport cost to encourage the disentanglement of source-specific features and source-agnostic features while promoting the perception of source-specific semantics

%In section \ref{4.2}, we derive source-specific subspaces by capturing the differences between each source and the barycenter space and design, we contrast the representations from the source-specific subspaces while ensuring their orthogonality to the barycenter representations. 

%This design encourages the disentanglement of source-specific features and source-agnostic features while promoting the perception of source-specific semantics.
\subsection{Multi-source Latent OT Objective}
\label{4.1}
%We aim to decompose the latent space into a source-agnostic unified space and $K$ source-specific subspaces. We seek the barycenter space for encoding the source-agnostic unified representations in the latent space, leveraging its inherent nature of capturing the averaging distribution of multiple source domains.
To learn non-trivial latent barycenters for unified source-agnostic features encoding, a key ingredient is the design of transport costs $C_k$, requiring appropriate modeling of the interrelations among multiple sources in the latent space. Here we propose the MLOT objective on the unit sphere to measure the distance of latent representations, in which the transport costs exploit the source-level contrastiveness and barycenter-anchored orthogonality.

Formally, given $K$ types of degraded images for training, the latent space is assumed to contain $K$ sources, in which the $k$-th source contains encoded features of the $k$-th type degraded images (Fig. \ref{pipeline}). The multi-source latent space can be written as $\mathcal Z=\cup_{k=1}^K\mathcal Z_k$, with distribution $\mathbb P_k$ for each source $\mathcal Z_k$. The barycenter space is denoted as $\mathcal Z_B$ with distribution $\mathbb Q$ as the barycenter distribution of $\mathbb P_{1:K}$. Given $z_k\in\mathcal Z_k$, we obtain its barycenter representation $z_k^B\in\mathcal Z_B$ via the barycenter map $T: \mathcal Z\rightarrow \mathcal Z_B$, \textit{i.e.,} $z_k^B=T(z_k)$. We denote source-specific subspaces as $\mathcal S_k$, with samples $s_k=z_k-z^B_k$ as the source-specific representations.

We first introduce the MLOT objective between the multiple source distributions $\mathbb P_k$ and the target barycenter distribution $\mathbb Q$ in the multi-source latent space:
\begin{align}
	\label{MLOT}
	\text{MLOT}_{C_k}(\mathbb{P}_k,&\mathbb{Q})\triangleq \nonumber\\\inf_{\pi\in\Pi(\mathbb P_k,\mathbb Q)}&\int_{\mathcal Z_k\times \mathcal Z_B}C_k(z_k,z^B_k)d\pi(z_k,z^B_k), 
\end{align}
with transport cost $C_k$ as
\begin{align}
	C_k(z_k,z^B_k)=\|z_k-z^B_k\|+\gamma(\mathcal L_{k}^{ctr}+\mathcal L_{k}^{ort}).\nonumber
\end{align}
Here $\mathcal L_{k}^{ctr}$ and $\mathcal L_{k}^{ort}$ are the terms to control the source-level contrastiveness and barycenter-anchored orthogonality. With the MLOT objective (\ref{MLOT}), we formulate the MLOT barycenter problem that seeks a barycenter space with distribution $\mathbb Q$ to encode multi-source unified representations, aggregating the multiple sources $\mathbb P_k$ with weights $\lambda_{1:K}$:
\begin{align}
	\mathcal L^*=\inf_{\mathbb Q\in\mathcal P(\mathcal Z_B)}\sum_{k=1}^{K}\lambda_k\text{MLOT}_{C_k}(\mathbb P_k, \mathbb Q). \label{mlot-bary}
\end{align}
\noindent\textbf{Source-level contrastiveness.} 
To learn source-specific representations with separated semantics and maximize mutual information between different source-specific subspaces, we introduce a source-level contrastive loss for the transport cost. Specifically, for $s_k\in\mathcal S_k$, we consider the representations in the same subspace $\mathcal S_k$ as positive samples $s_{k}^+$. The negative samples $s_k^{-}$ are representations from other source-specific subspaces $\mathcal S_i~(i\not=k, i\in\bar K)$. By letting $s_k$ attract positive samples and repel the negative ones, the source-level contrastive loss for the $k$-th source can be defined as
\begin{small}
	\begin{align*}
		\mathcal L_{k}^{ctr}\triangleq-\log\frac{\displaystyle\sum_{s_{k}^+\in\mathcal S_k}\exp(\langle s_k,s_{k}^+\rangle/\tau)}{\displaystyle\sum_{s_{k}^+\in\mathcal S_k}\exp(\langle s_k,s_{k}^+\rangle/\tau)+\sum_{s_{k}^-\in \mathcal{S}_i}\exp(\langle s_k,s_k^{-}\rangle/\tau)}, 
	\end{align*}
\end{small}

\noindent where $\tau$ is the temperature hyper-parameter.  In practice, the contrastive loss is incorporated in the transport cost of the MLOT objective, and $s_k$, along with its positive/negative samples, can be sampled over mini-batches. 

\quad\\
\noindent\textbf{Barycenter-anchored orthogonality.}  
To promote the decomposition of source-agnostic  and source-specific features in the  multi-source latent space, we define the barycenter-anchored orthogonal loss for the $k$-th source as follows:
\begin{align*}
	\mathcal L_k^{ort}\triangleq\sum_{s_j\in\mathcal S_j}|\langle z^B_k, s_j \rangle|,
\end{align*}
where $\mathcal S_j$ with $j\in\bar K$ covers all the source-specific subspaces. This orthogonal loss ensures the orthogonality between the barycenter space and source-specific subspaces. In this sense, the established barycenter space encodes compact representations that capture shared information across sources while discarding source-specific nuisance factors. 

%In section \ref{4.2} below, we show that the MLOT barycenter problem can be solved in a maximin manner and derive the barycenter map with its error bounds.
\subsection{MLOT Barycenter Map}
\label{4.2}
For convenience, we introduce the following functional:
\begin{align}
	\mathcal L(f_{1:K})\triangleq\sum_{k=1}^K\lambda_k\int_{\mathcal Z_k} f_k^{C_k}(z_k)d\mathbb P_k(z_k),
	\label{func1}
\end{align}
with the $C_k$-transform of $f_k$:
\begin{align}
	f_k^{C_k}(z_k)=\inf_{z^B_k\in\mathcal Z_B}\big[C_k(z_k,z^B_k)-f_k(z_k^B)\big]. \nonumber
\end{align}
Given the challenge of directly solving the MLOT problem (\ref{mlot-bary}), we present its dual reformulation in Theorem \ref{the-bary} below. This theorem enables us to compute the barycenters in a maximin optimization manner if the potentials $f_{1:K}\in \mathcal C(\mathcal Z_B)^K$ satisfy the congruence condition $\sum_{k=1}^K\lambda_kf_k\equiv0$.
\begin{theorem}[\normalfont{Dual reformulation for MLOT barycenter problem (\ref{mlot-bary})}]The minimum objective value $\mathcal{L}^*$ of the MLOT barycenter problem (\ref{mlot-bary}) can be expressed as 	\label{the-bary}	
	\begin{align}
		\mathcal L^*=\sup\limits_{\substack{\sum_k\!\lambda_k f_k = 0;\\f_{1},\ldots,f_k\in\mathcal C(\mathcal Z_B)}}\mathcal L(f_{1:K}).
	\end{align}
\end{theorem}
Now we aim to seek the barycenter map $T:\mathcal Z\rightarrow \mathcal Z_B$. By substituting  the optimization over target $z^B_k\in\mathcal Z_B$ with an equivalent optimization over the barycenter map of interest $T$ (guaranteed by Rockafellar interchange theorem \cite{rockafellar1976integral}, Theorem 3A), we can reformulate Eq. (\ref{func1}) as
\begin{align}
	\mathcal L(f_{1:K})&=\inf_{T}\bigg\{\sum_{k=1}^K\lambda_k\int_{\mathcal Z_k} \big[C_k(z_k,T(z_k))\nonumber\\&-f_k(T(z_k))\big]d\mathbb P_k(z_k)\bigg\}.
	\label{func2}
\end{align}

We denote the expression under $\inf$ in (\ref{func2})  by $\mathcal F(f_{1:K},T)$. Then the objective can be written as the maximin form
\begin{align}
	\mathcal L^*=\!\!\!\!\!\!\sup_{\substack{\sum_k\!\lambda_k f_k = 0;\\f_{1},\ldots,f_k\in\mathcal C(\mathcal Z_B)}}\inf_{\substack{\quad\\T:\mathcal Z\rightarrow \mathcal Z_B}}\mathcal F(f_{1:K},T).
	\label{maximin}
\end{align}
\textbf{Error Bounds.} We answer the question of how close the estimated map $\widehat T$ is to the true barycenter map $T^*$ that transports $\mathbb P_k$ and the barycenter $\mathbb Q^*$. We establish the error bound for the estimated barycenter map in Theorem \ref{error}, which demonstrates that for the pair $(\widehat f_{1:K},\widehat T)$ that solves the optimization problem (\ref{maximin}), the recovered map $\widehat T$ is close to the true barycenter map $T^*$.
\begin{theorem}[\normalfont{Error analysis via duality gaps for the recovered maps}] Let $C_k$ be the MLOT transport costs. Assume that the maps $z^B_k\to C_k(z_k,z^B_k)-\widehat f_k(z^B_k)$ are $\beta$-strongly convex for $z_k\in\mathcal Z_k$, $k\in\bar K$. Consider the duality gaps for an approximate solution $(\widehat f_{1:K},\widehat T)$ of {\normalfont(\ref{maximin})}:
	\begin{align}
		\mathcal E_1(\widehat{f}_{1:K}, \widehat{T}) &\triangleq \mathcal F(\widehat{f}_{1:K}, \widehat T) - \mathcal L(\widehat{f}_{1:K}); \label{eq:deltagap1} \\
		\mathcal E_2(\widehat{f}_{1:K}) &\triangleq \mathcal L^* - \mathcal L(\widehat{f}_{1:K}), \label{eq:deltagap2}
	\end{align}
	which are the errors of solving the inner $\inf$ and outer $\sup$ problems in (\ref{maximin}). Then the following inequality holds: \label{error}
	\begin{align}
		\sum_{k=1}^{K}\lambda_{k}\mathbb{W}_{2}^{2}\left(\widehat T_{\#}\mathbb P_k,T^{*}_\#\mathbb P_k\right)\leq\frac{2}{\beta}(\mathcal E_1+\mathcal E_2).\nonumber
	\end{align}
\end{theorem}
\subsection{Parameterization and Optimization Algorithm}
\label{4.3}
To tackle the formulated MLOT barycenter problem (\ref{maximin}), we parameterize the barycenter map $T$ and potentials $f_{k}$ ($k\in\bar K)$ with NNs $T_\theta$ and $f_{\omega_k}$ (see implementation details). The maximin optimization objective can be written as
\begin{align}
	\mathcal F(\omega_{1:K},\theta)\!=\!\sum_{k=1}^K\lambda_k\!\!\mathop{\mathbb E}_{z_k\sim\mathbb P_k}\!\big[C_k(z_k,T_\theta(z_k))-f_{\omega_k}(T_\theta(z_k))\big].\nonumber
\end{align}
The weights $\lambda_k$ are set to the portion of the number of training samples for each source. We introduce a congruence penalty  defined as $\rho(\omega_{1:K})=\|\sum_{i=1}^K\lambda_if_{\omega_i}(z^B_k)\|^2$  to ensure the congruence condition. Finally, we adversarially train networks $T_\theta$ and $f_{\omega_k}$ by minimizing and maximizing $\mathcal F(\omega_{1:K},\theta)$, respectively, while penalizing the congruence condition. This process boils down to
\begin{align}
	\max_{\omega_{1:K}}\min_\theta\{\mathcal F(\omega_{1:K},\theta)-\rho(\omega_{1:K})\},
\end{align} 
where we estimate the expectation using mini-batch data in each training step. 
The algorithm for training $T_\theta$ and $f_{\omega_{1:K}}$ is detailed in the \textbf{supplementary material}.

Besides the training of the barycenter map $T_\theta$, we adopt end-to-end pairwise training using $L_1$ loss for the overall restoration network without pretraining any individual component. At test time, the learned barycenter map transforms the encoded features of degraded images into the barycenter ones, which are aggregated with the source-specific representations and decoded into clear images (Fig. \ref{pipeline}).

\section{Experiments}
\label{exp}
We evaluate BaryIR under both the all-in-one and task-specific configurations on benchmark datasets across multiple restoration tasks. We also evaluate its generalization performance on unseen real-world scenarios and unseen degradation levels. The best and second-best results are \textbf{highlighted} and \underline{underlined}. The \textbf{supplementary material} provides the implementation details, ablation study for the weights $\lambda_{1:K}$, dataset details, task-specific restoration results, evaluation metrics, and further model analyses. 

\subsection{All-in-One Restoration Results}
\begin{table*}[!htbp]
	\centering
	\vspace{-0.1cm}
	\caption{The \textbf{All-in-One three-degradation} results. The metrics are reported as PSNR($\uparrow$)/SSIM($\uparrow$)/LPIPS($\downarrow$)/FID($\downarrow$). }
	\label{air3}
	\vspace{-0.2cm}
	\setlength{\tabcolsep}{3pt}
	\renewcommand{\arraystretch}{1.2}
	\resizebox{\textwidth}{!}{
		\begin{tabular}{lcccccc}
			\toprule
			\multirow{2}{*}{Method} 
			&\textit{Dehazing} & \textit{Deraining} & \multicolumn{3}{c}{\textit{Denoising}} 
			& \multirow{2}{*}{Average}  \\  \cmidrule(lr){2-2} \cmidrule(lr){3-3} \cmidrule(lr){4-6} 
			& SOTS &Rain100L & BSD68\textsubscript{$\sigma$=15} & BSD68\textsubscript{$\sigma$=25} & BSD68\textsubscript{$\sigma$=50} &  \\\midrule
			
			MPRNet \cite{Zamir2021MPRNet}    & 25.43/0.956/0.038/28.15 &  33.66/0.955/0.057/33.65  &  33.50/0.925/0.084/52.87&  30.89/0.880/0.127/79.53&   27.48/0.778/0.201/121.9 & 30.19/0.899/0.101/63.23  \\
			Restormer \cite{Zamir2021Restormer}   & 29.92/0.970/0.035/22.29 &  35.64/0.971/0.036/33.97  &  33.81/0.932/0.078/42.61&  31.00/0.880/0.113/74.62&   27.85/0.792/0.198/117.6 & 31.62/0.909/0.092/58.22 \\
			IR-SDE \cite{luo2023image}   & 29.35/0.961/0.029/19.80 &  34.87/0.958/0.031/30.36  &  32.89/0.903/0.068/35.51	&  30.56/0.861/0.107/68.15&   27.22/0.769/0.195/107.6 & 30.98   0.890/0.086/52.29\\\midrule
			
			%			AirNet \cite{li2022all} & 27.94/0.962/0.030/26.71&34.80/0.962/0.054/33.28&33.92/\underline{0.933}/0.066/45.67&31.26/0.888/0.110/72.74&28.00/0.797/0.194/106.6&31.18/0.908/0.091/57.00\\
			%			IDR \cite{zhang2023ingredient}&28.68/0.968/0.018/18.55&35.99/0.968/0.025/23.65&33.89/0.932/0.063/39.13&31.33/0.889/0.103/63.79&28.02/0.798/0.187/102.6&31.64/0.911/0.079/49.54\\
			PromptIR \cite{potlapalli2023promptir}& \underline{30.58}/\underline{0.974}/0.012/13.23&36.37/0.972/0.019/16.78&33.97/\underline{0.933}/0.046/27.54&31.29/0.888/0.090/53.69&28.06/0.798/0.179/95.84&32.05/\underline{0.913}/0.069/41.42\\
			DA-CLIP \cite{luocontrolling}&30.12/0.972/\underline{0.009}/\underline{8.952}&35.92/0.972/0.015/13.73&33.86/0.925/0.045/25.27&31.06/0.865/0.082/48.64&27.55/0.778/0.168/89.28&31.70/0.901/0.063/37.17\\
			RCOT \cite{tang2024residualconditioned}  & 30.32/0.973/\underline{0.009}/10.52&37.25/0.974/0.015/\underline{12.25}&33.86/0.932/0.048/30.12&31.20/0.886/0.086/57.25&28.03/0.797/\underline{0.162}/\underline{87.69}&32.13/0.912/0.065/39.57\\
			DiffUIR \cite{zheng2024selective}  &30.18/0.973/0.010/10.23&36.78/0.973/\underline{0.013}/12.62&33.94/0.932/\underline{0.044}/\underline{24.95}&31.26/0.887/\underline{0.080}/\underline{46.12}&28.04/0.797/0.164/88.10&32.04/0.912/\underline{0.062}/\underline{36.40}\\
			InstructIR \cite{conde2024instructir} &30.22/0.959/0.012/14.56&\underline{37.98}/\underline{0.978}/0.021/20.52&\underline{34.15}/\underline{0.933}/0.051/33.45&\underline{31.52}/\underline{0.890}/0.088/55.76&\textbf{28.30}/\textbf{0.804}/0.175/98.19&\underline{32.43}/\underline{0.913}/0.070/44.50\\
			\midrule
			BaryIR & \textbf{31.33}/\textbf{0.980}/\textbf{0.007}/\textbf{4.523}&\textbf{38.95}/\textbf{0.984}/\textbf{0.008}/\textbf{5.739}& \textbf{34.16}/\textbf{0.935}/\textbf{0.038}/\textbf{22.69}&
			\textbf{31.54}/\textbf{0.892}/\textbf{0.075}/\textbf{40.11} & \underline{28.25}/\underline{0.802}/\textbf{0.158}/\textbf{82.63} &\textbf{32.85}/\textbf{0.919}/\textbf{0.057}/\textbf{31.14}\\
			\bottomrule
	\end{tabular}}
	\vspace{0.1mm}
	\caption{The \textbf{All-in-One five-degradation} results. The metrics are reported as PSNR($\uparrow$)/SSIM($\uparrow$)/LPIPS($\downarrow$)/FID($\downarrow$). }
	\label{air5}
	\vspace{-0.2cm}
	\setlength{\tabcolsep}{3pt}
	\renewcommand{\arraystretch}{1.2}
	\resizebox{\textwidth}{!}{
		\begin{tabular}{lcccccc}
			\toprule
			\multirow{2}{*}{Method} 
			&\textit{Dehazing} & \textit{Deraining} & 
			\textit{Denoising}&\textit{Deblurring}&\textit{Low-light}
			& \multirow{2}{*}{Average}  \\  \cmidrule(lr){2-2} \cmidrule(lr){3-3} \cmidrule(lr){4-4} \cmidrule(lr){5-5} \cmidrule(lr){6-6} 
			& SOTS &Rain100L &  BSD68\textsubscript{$\sigma$=25} & GoPro& LOL-v1 &  \\
			\midrule
			
			MPRNet \cite{Zamir2021MPRNet}   & 24.28/0.931/0.061/43.55 &  33.12/0.927/0.064/57.84  &  30.18/0.846/0.112/83.47&  25.98/0.786/0.179/55.95&   18.98/0.776/0.115/103.5 & 26.51/0.853/0.106/68.86  \\
			Restormer \cite{Zamir2021Restormer}   & 24.09/0.927/0.065/41.76 &  34.81/0.971/0.045/49.18  &  30.78/0.876/0.095/72.95&  27.22/0.829/0.174/56.10&   20.41/0.806/0.109/107.7 & 27.46/0.881/0.098/65.54 \\
			IR-SDE \cite{luo2023image}   & 24.56/0.940/0.047/29.89 &  34.12/0.951/0.040/43.95  &  30.89/0.865/0.089/62.16
			&  26.34/0.800/0.162/48.77&   20.07/0.780/0.102/86.13 & 27.20/0.867/0.088/54.18\\
			\midrule
			%			AirNet \cite{li2022all} &21.04/0.884/0.077/62.52 & 32.98/0.951/0.058/50.12 & 30.91/0.882/0.102/78.12 & 24.35/0.781/0.189/66.13 & 18.18/0.735/0.122/116.9&25.49/0.847/0.110/74.76\\
			%			IDR \cite{zhang2023ingredient}&25.24/0.943/0.052/33.25 & 35.63/0.965/0.043/45.62 & \textbf{31.60}/0.887/0.092/66.24 & 27.87/0.846/0.178/40.83 & 21.34/0.826/0.108/100.6&28.33/0.893/0.095/57.31\\
			PromptIR \cite{potlapalli2023promptir} &\underline{30.41}/\underline{0.972}/0.017/20.12 & 36.17/0.970/0.024/22.53 & 31.20/0.885/0.097/66.91 & 27.93/0.851/0.155/29.52 & \underline{22.89}/0.829/0.098/70.32&29.72/0.901/0.078/41.88\\
			DA-CLIP \cite{luocontrolling}&29.78/0.968/0.014/15.26&35.65/0.962/0.022/22.24&30.93/0.885/\underline{0.089}/\underline{54.12}&27.31/0.838/\underline{0.143}/\underline{23.34}&21.66/0.828/\underline{0.095}/\underline{55.81}&29.07/0.896/\underline{0.073}/\underline{34.15}\\
			RCOT \cite{tang2024residualconditioned}& 30.26/0.971/0.016/16.74&\underline{36.88}/\underline{0.975}/0.024/\underline{19.67}& 31.05/0.882/0.099/62.12&28.12/0.862/0.155/21.56& 22.76/0.830/0.097/61.24&29.81/0.904/0.078/36.26\\
			DiffUIR \cite{zheng2024selective}  &29.47/0.965/\underline{0.013}/\underline{15.01}&35.98/0.968/\underline{0.020}/20.45&31.02/0.885/0.093/58.17&27.50/0.845/0.147/26.65&22.32/0.826/0.097/60.21&29.25/0.898/0.074/36.10\\
			InstructIR \cite{conde2024instructir} &27.10/0.956/0.015/16.28&36.84/0.973/0.025/23.86&31.40/\underline{0.890}/0.102/63.69&\textbf{29.40}/\underline{0.886}/0.158/35.29&\underline{23.00}/\underline{0.836}/0.102/65.86&29.55/0.908/0.081/41.00\\
			\midrule
			BaryIR& \textbf{31.12}/\textbf{0.976}/\textbf{0.010}/\textbf{6.552}&\textbf{38.05}/\textbf{0.981}/\textbf{0.015}/\textbf{10.64}& \textbf{31.43}/\textbf{0.891}/\textbf{0.086}/\textbf{43.22}&\underline{29.30}/\textbf{0.888}/\textbf{0.141}/\textbf{15.47} & \textbf{23.38}/\textbf{0.852}/\textbf{0.092}/\textbf{51.48} &\textbf{30.66}/\textbf{0.918}/\textbf{0.069}/\textbf{25.47}\\
			\bottomrule
			\vspace{0.5mm}
	\end{tabular}}
	\setlength\tabcolsep{1pt}
	\renewcommand{\arraystretch}{0.7} % Default value: 1
	\centering
	\begin{tabular}{cccccc}
		\includegraphics[width=0.16\linewidth]{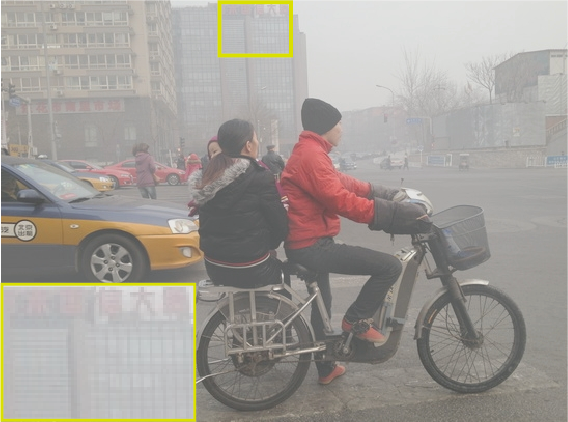}&
		\includegraphics[width=0.16\linewidth]{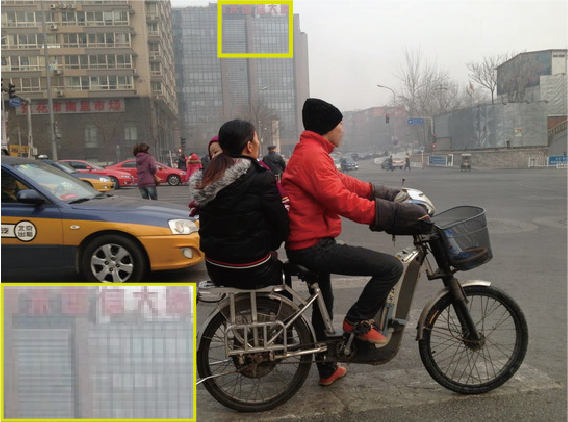}&
		\includegraphics[width=0.16\linewidth]{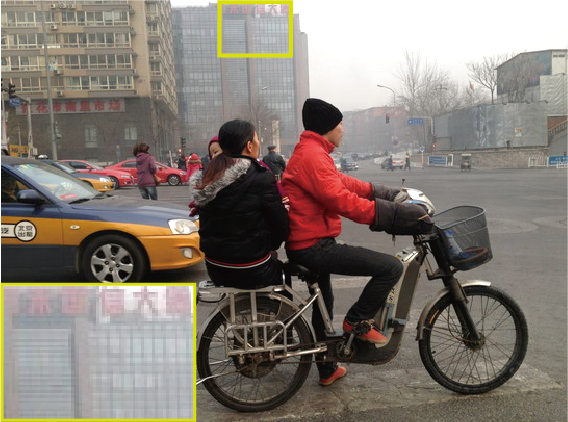}&
		\includegraphics[width=0.16\linewidth]{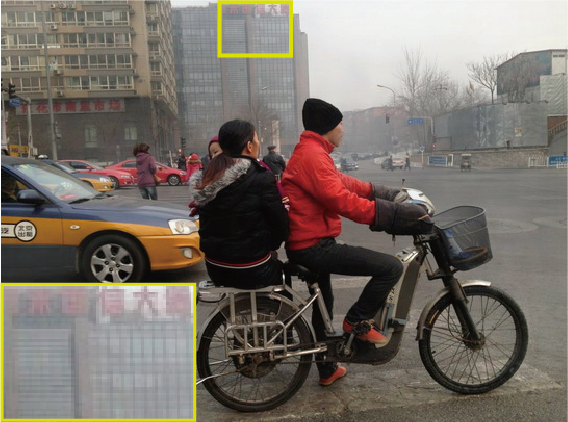}&
		\includegraphics[width=0.16\linewidth]{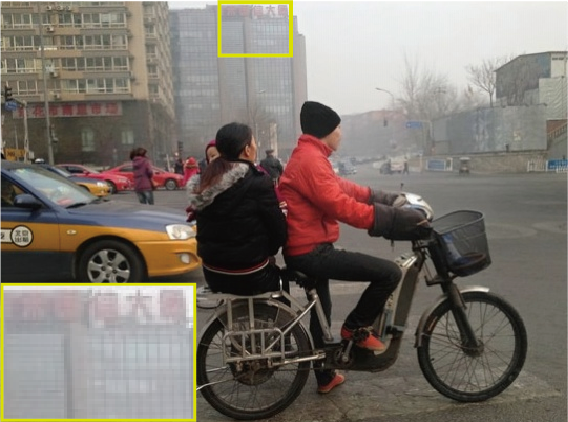}&
		\includegraphics[width=0.16\linewidth]{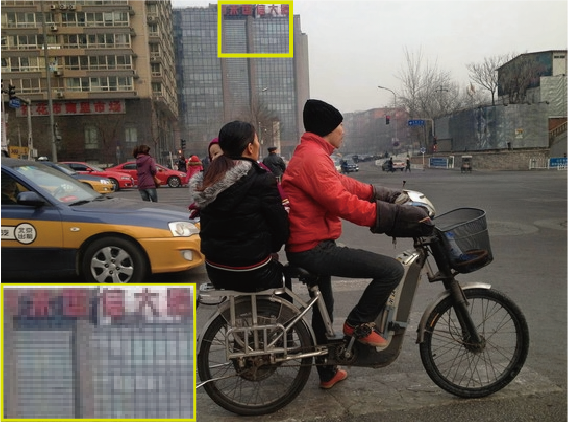}\\
		\includegraphics[width=0.16\linewidth]{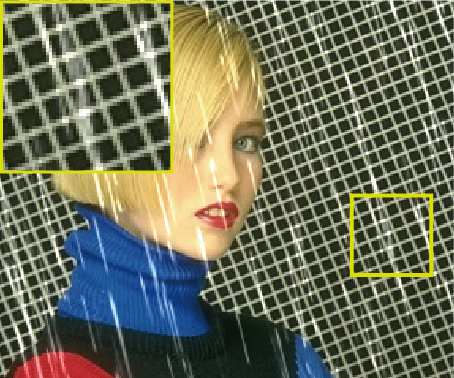}&
		\includegraphics[width=0.16\linewidth]{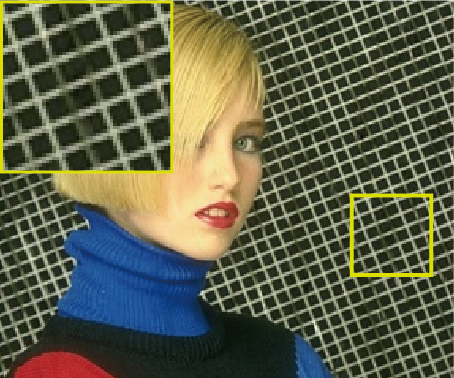}&
		\includegraphics[width=0.16\linewidth]{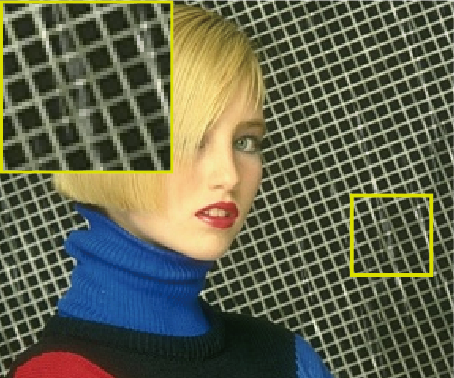}&
		\includegraphics[width=0.16\linewidth]{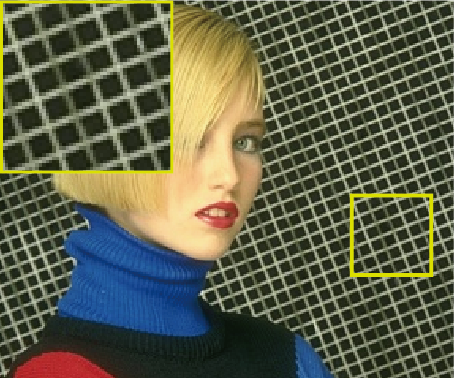}&
		\includegraphics[width=0.16\linewidth]{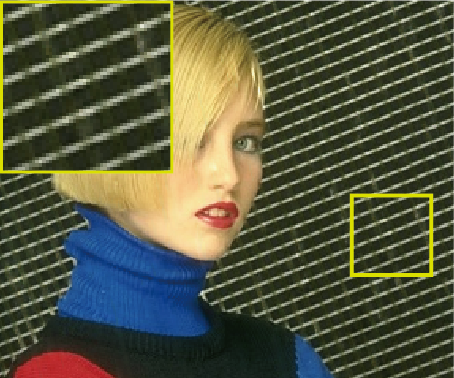}&
		\includegraphics[width=0.16\linewidth]{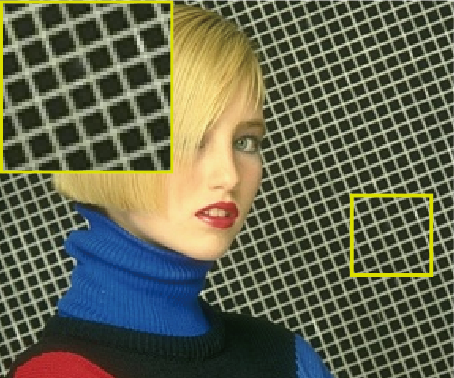}\\
		\includegraphics[width=0.16\linewidth]{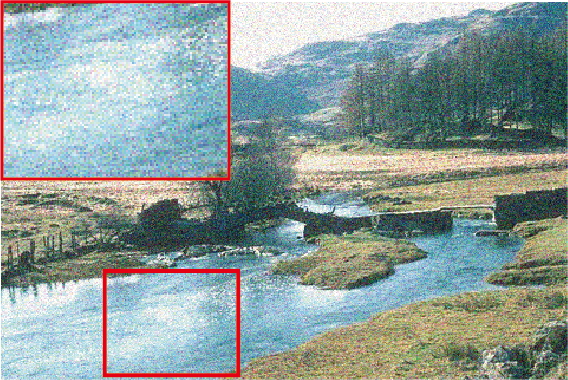}&
		\includegraphics[width=0.16\linewidth]{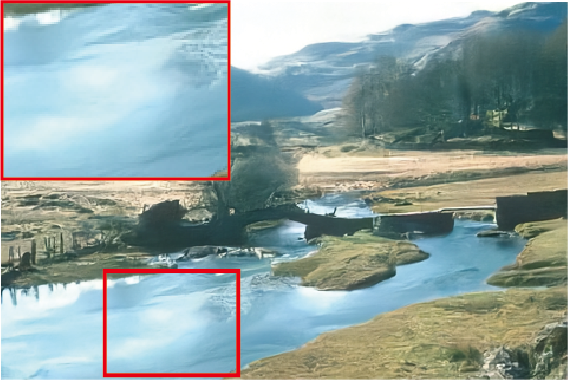}&
		\includegraphics[width=0.16\linewidth]{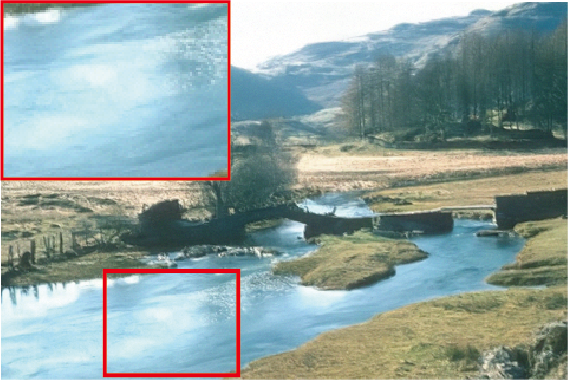}&
		\includegraphics[width=0.16\linewidth]{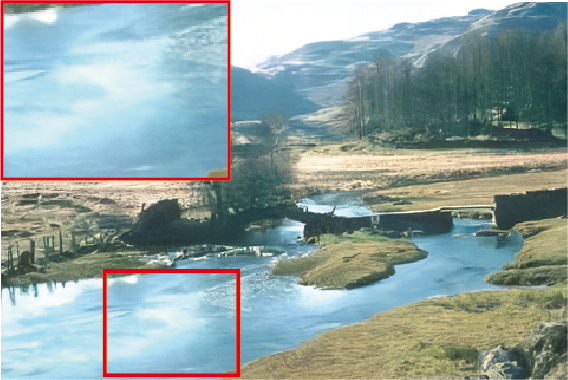}&
		\includegraphics[width=0.16\linewidth]{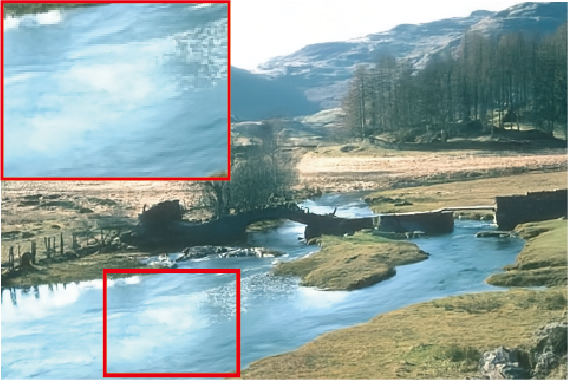}&
		\includegraphics[width=0.16\linewidth]{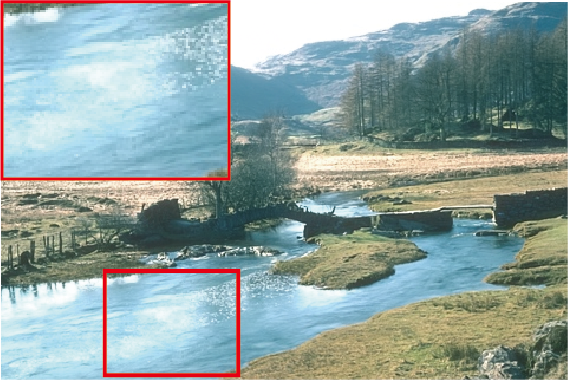}\\
		\includegraphics[width=0.16\linewidth]{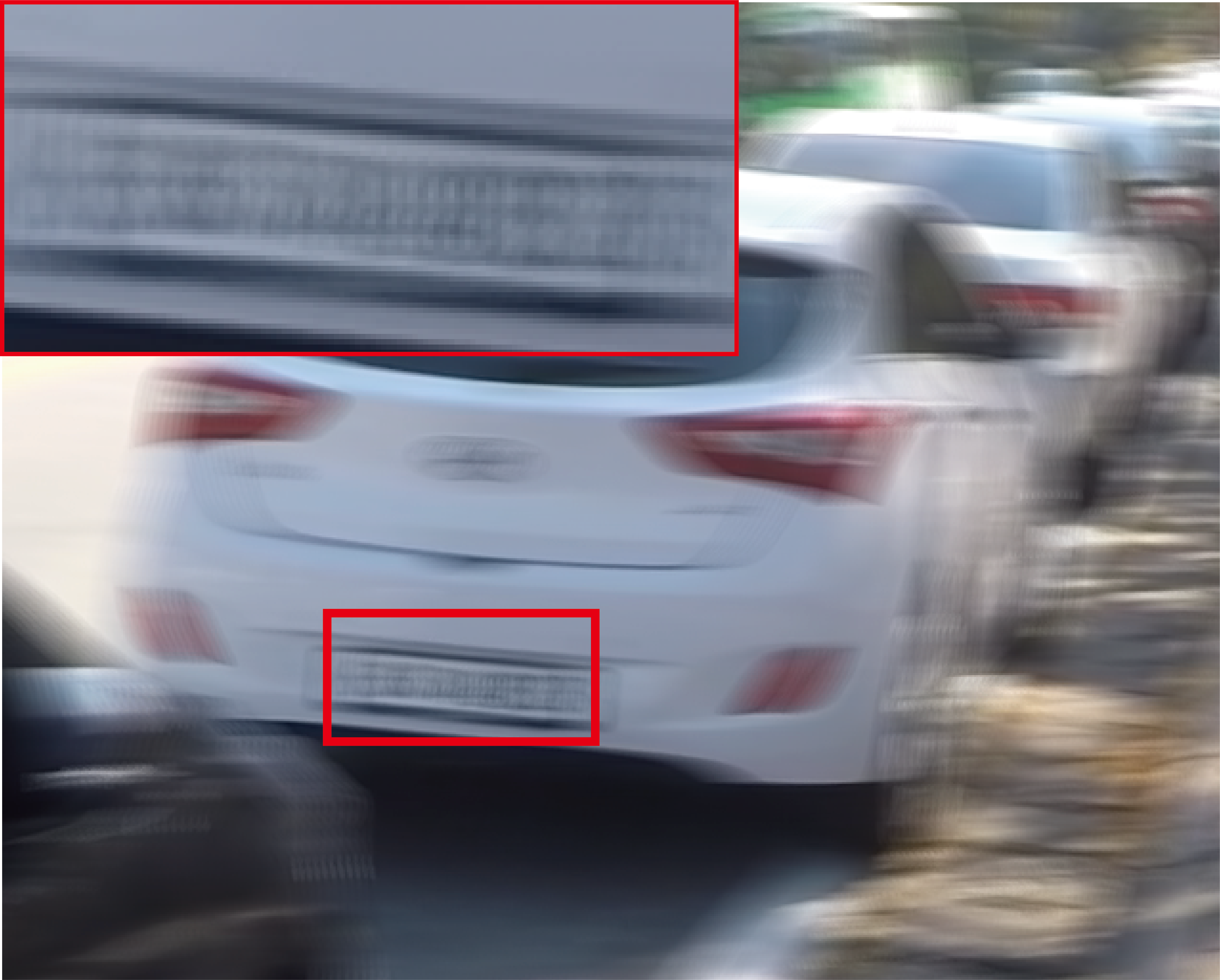}&
		\includegraphics[width=0.16\linewidth]{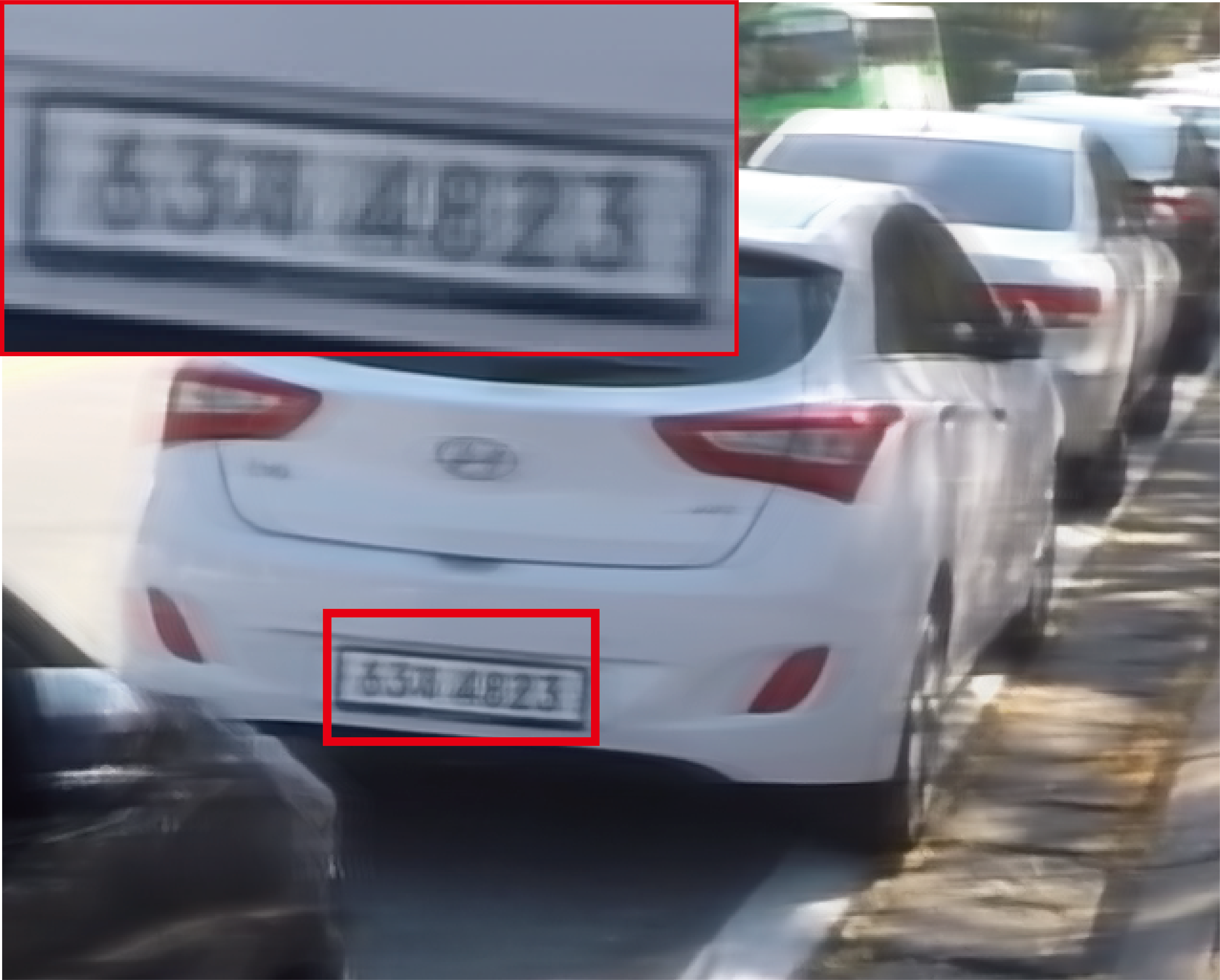}&
		\includegraphics[width=0.16\linewidth]{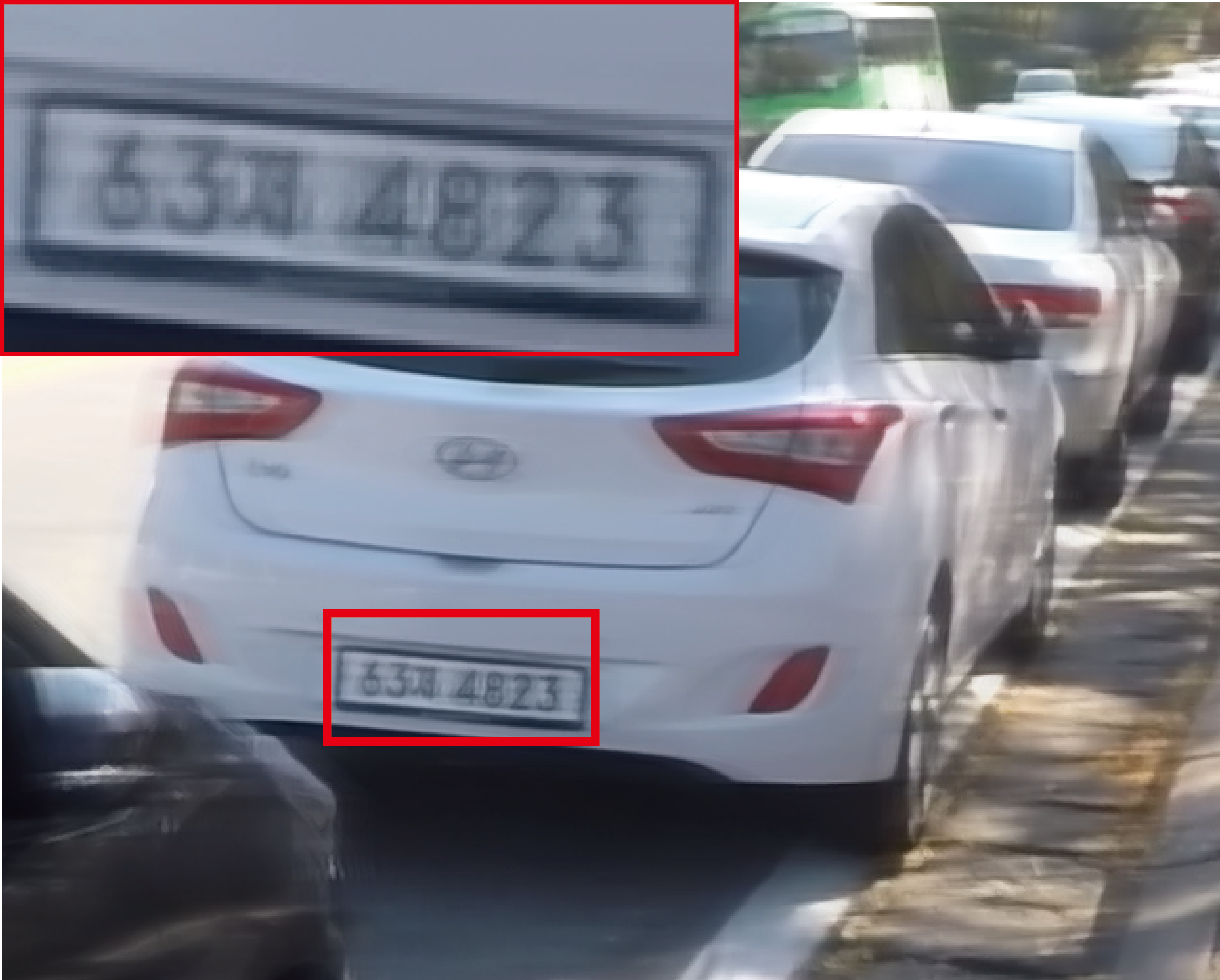}&
		\includegraphics[width=0.16\linewidth]{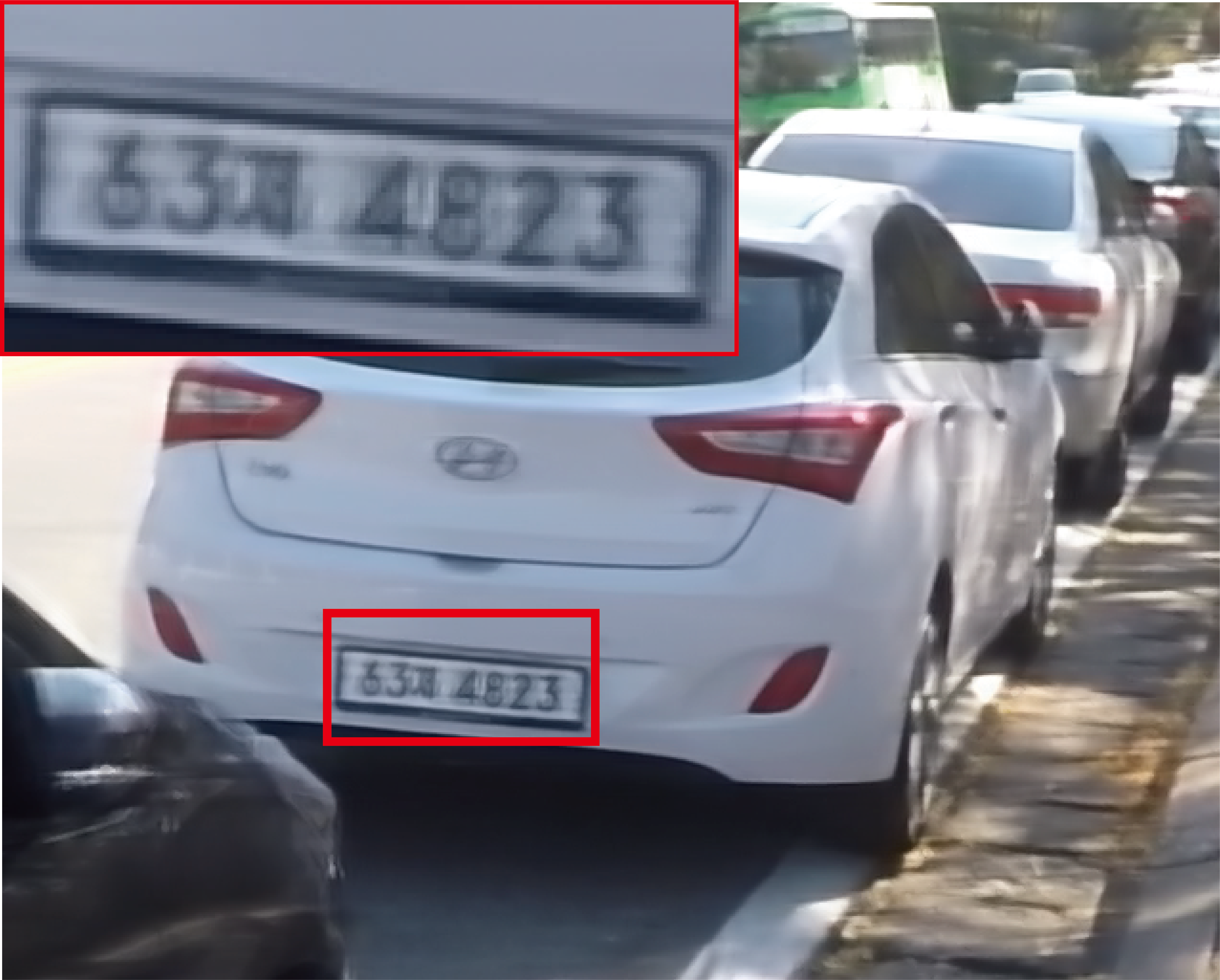}&
		\includegraphics[width=0.16\linewidth]{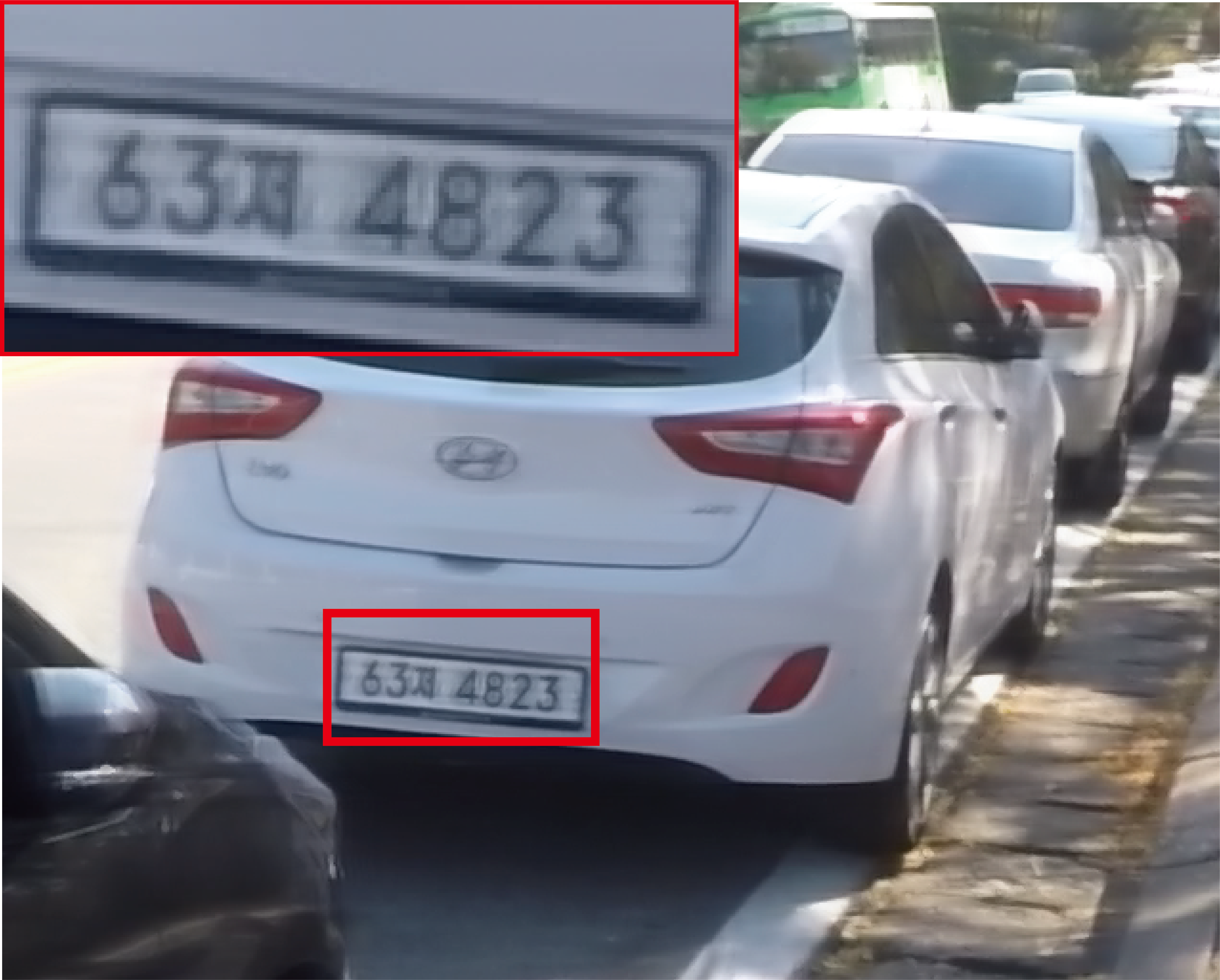}&
		\includegraphics[width=0.16\linewidth]{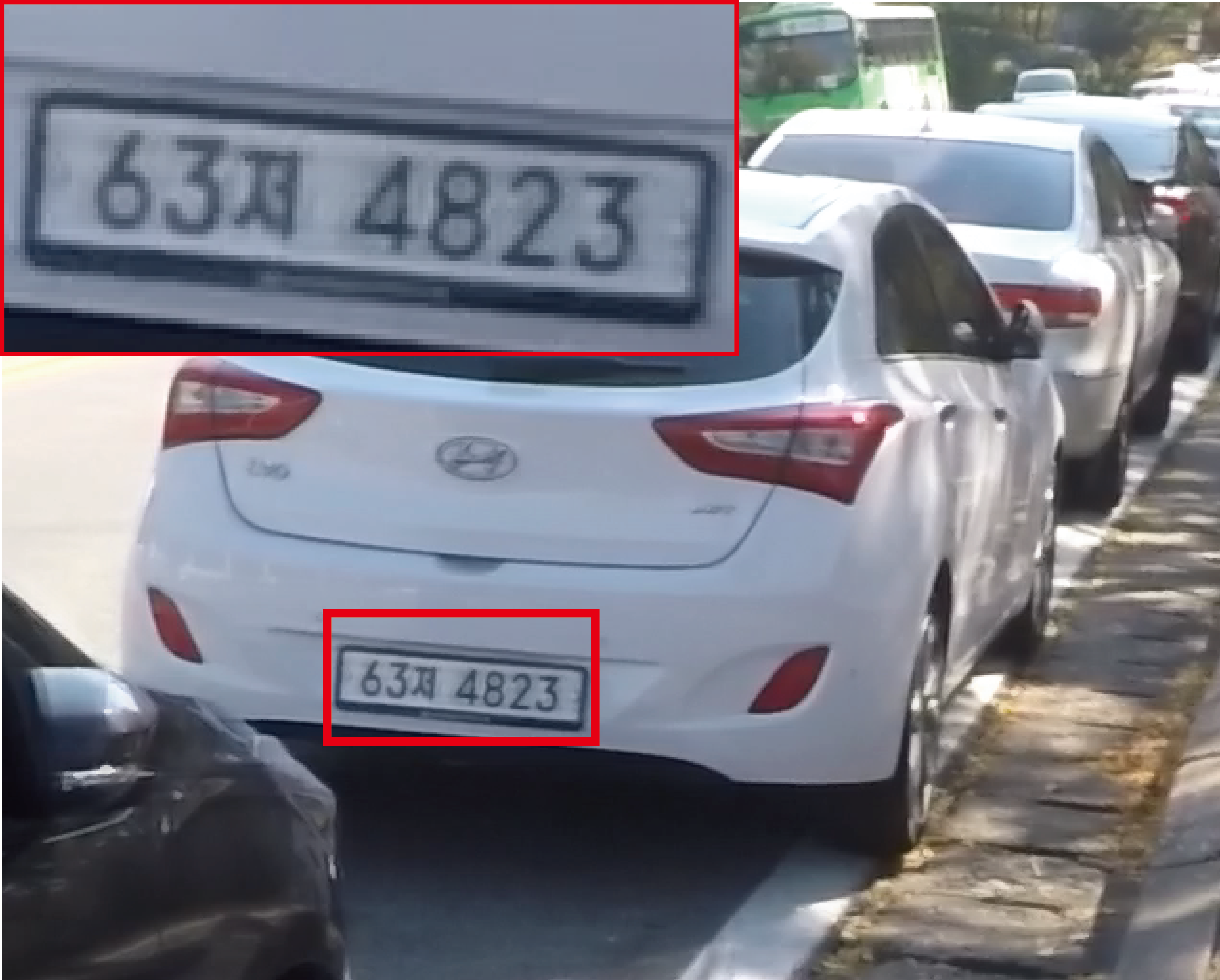}\\
		\includegraphics[width=0.16\linewidth]{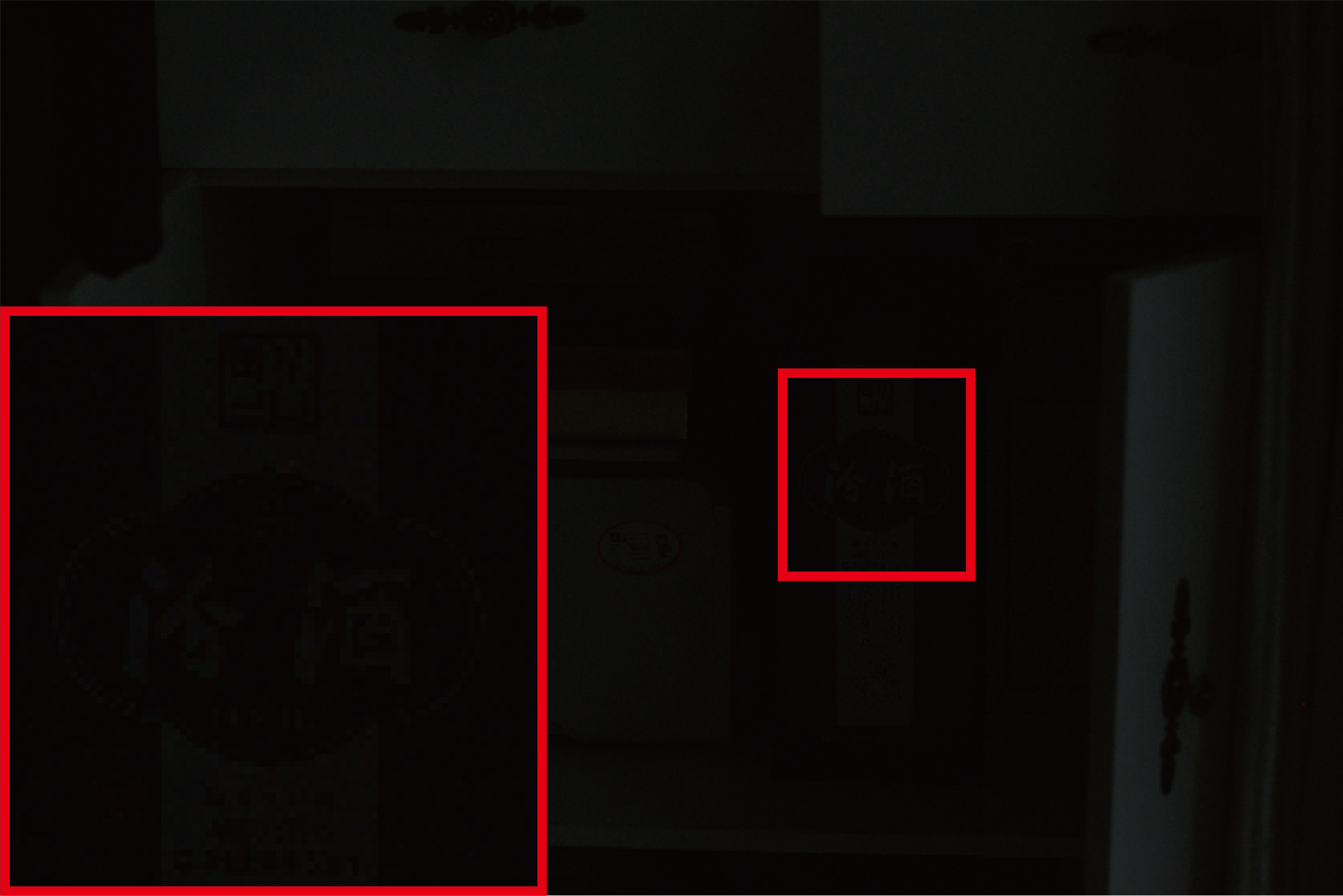}&
		\includegraphics[width=0.16\linewidth]{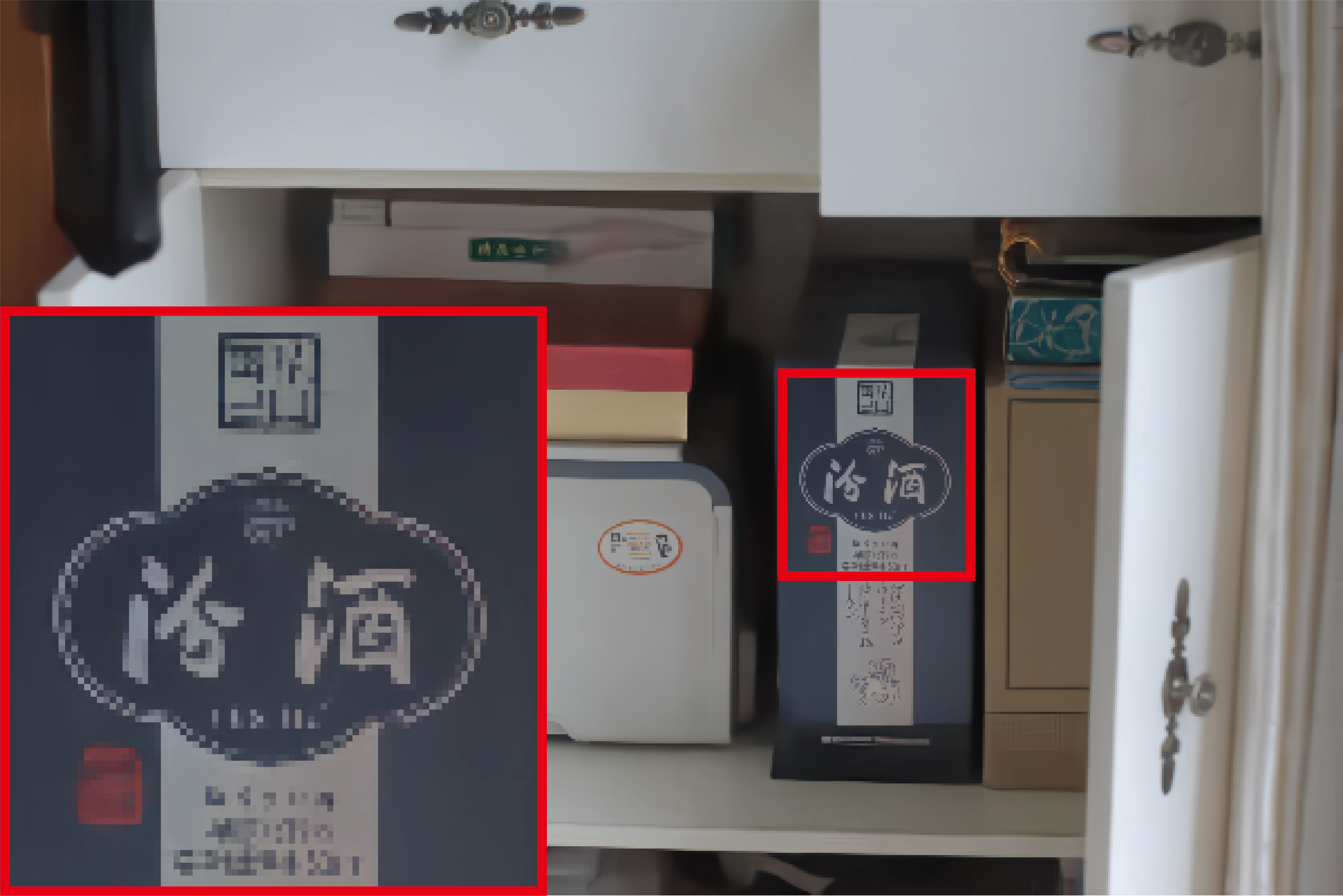}&
		\includegraphics[width=0.16\linewidth]{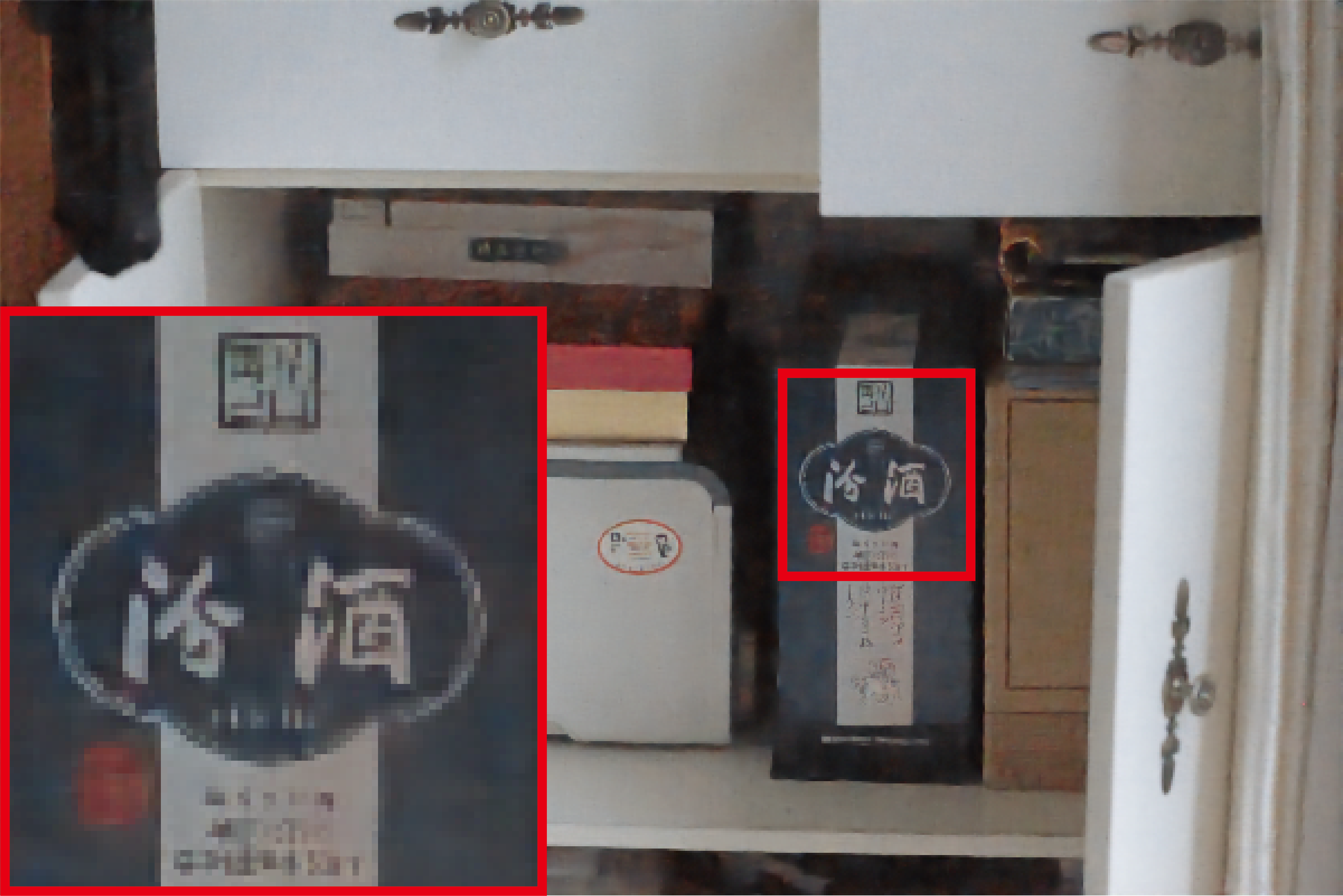}&
		\includegraphics[width=0.16\linewidth]{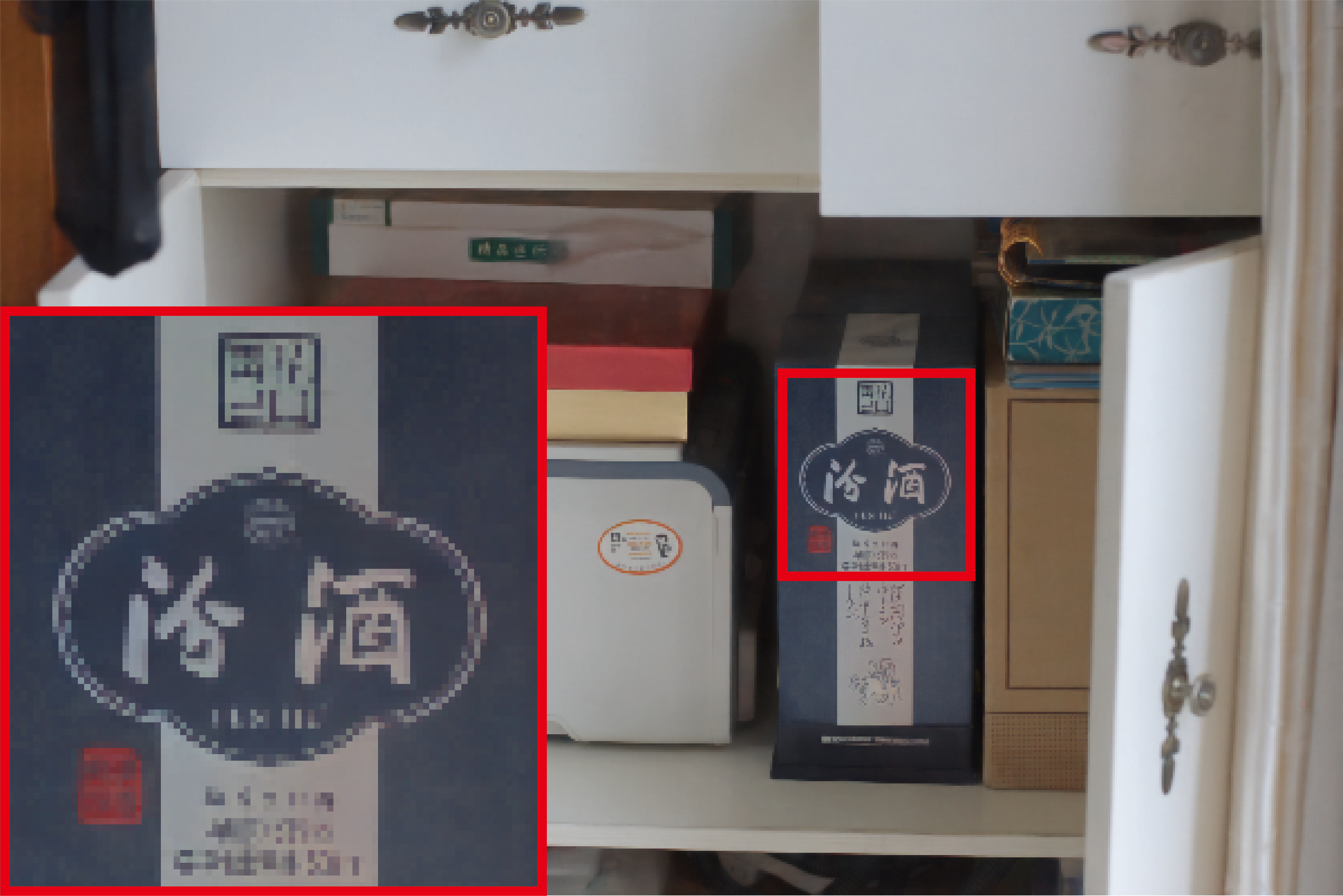}&
		\includegraphics[width=0.16\linewidth]{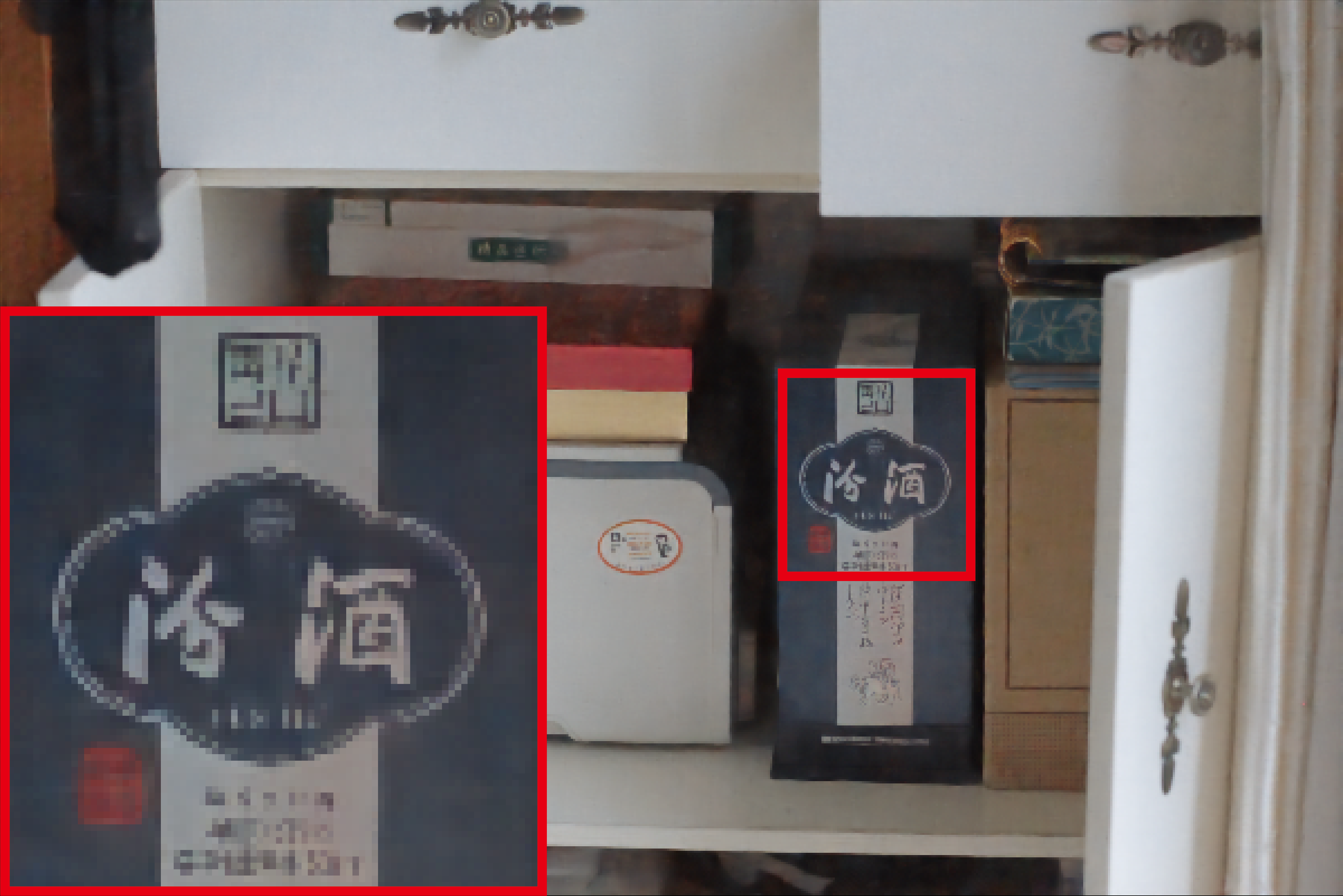}&
		\includegraphics[width=0.16\linewidth]{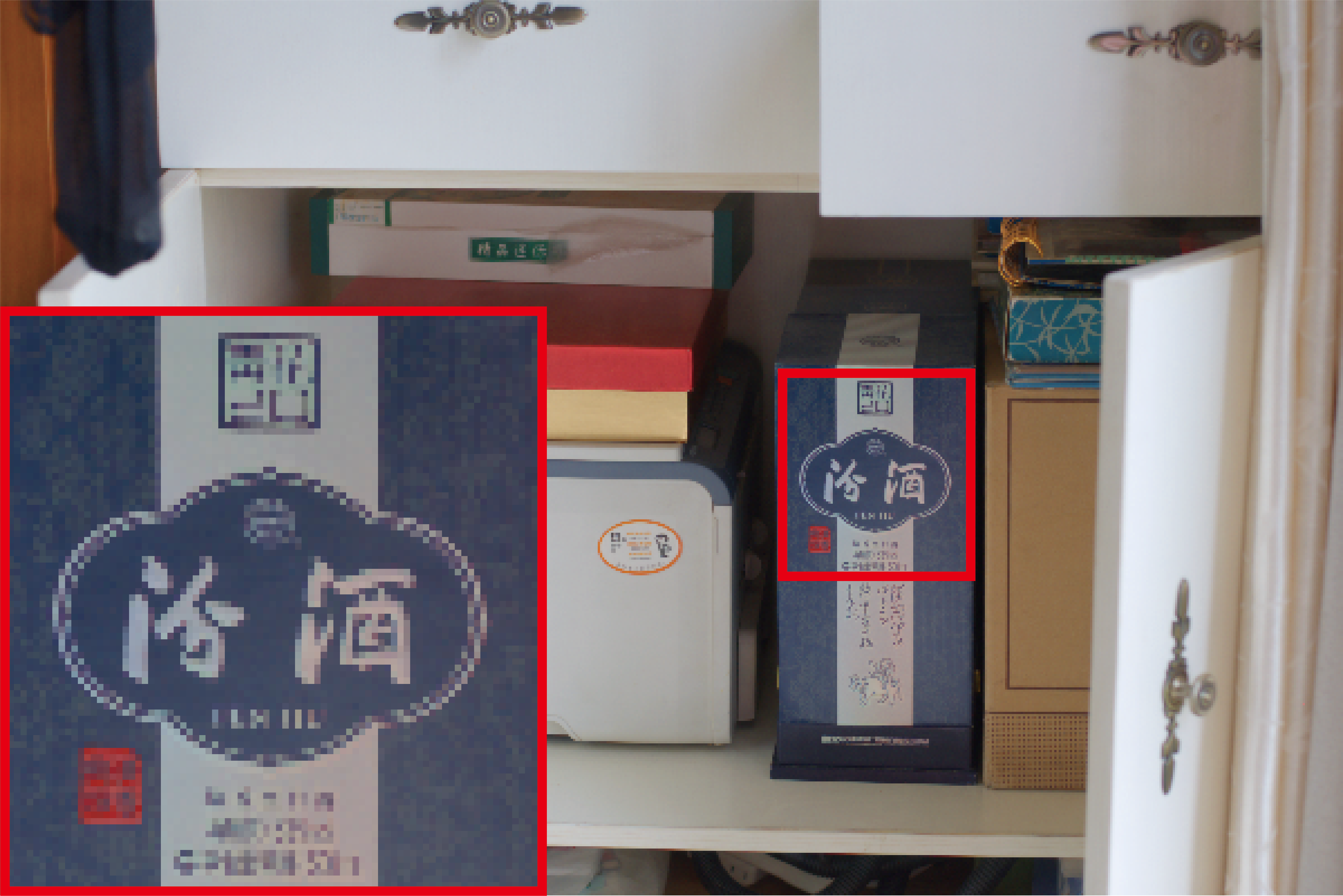}\\
		Degraded&PromptIR&DA-CLIP&RCOT&InstructIR&BaryIR\\
	\end{tabular}
	\vspace{-0.1cm}
	\captionsetup{type=figure}
	\caption{Visual comparison of five-degradation All-in-One results. BaryIR restores sharp images with fine-grained details.}
	\label{vair5}
\end{table*}
For the All-in-One configuration, we compare BaryIR with SOTA methods including three general restorers, \textit{i.e.,} MPRNet \cite{Zamir2021MPRNet}, Restormer \cite{Zamir2021Restormer}, IR-SDE \cite{luo2023image}; and five recent All-in-One models, \textit{i.e.,} PromptIR \cite{potlapalli2023promptir}, DA-CLIP \cite{luocontrolling}, RCOT \cite{tang2024residualconditioned}, DiffUIR \cite{zheng2024selective}, and InstructIR \cite{conde2024instructir}. Following the standard setting of prior works \cite{conde2024instructir}, \cite{potlapalli2023promptir}, we evaluate on the three-degradation and five-degradation benchmarks.

\textbf{Three degradations.} The first comparison is conducted across three restoration tasks: dehazing, deraining, and denoising at noise levels $\sigma\in\{15,25,50\}$. Tab. \ref{air3} reports the quantitative results, showing that BaryIR offers consistent performance gains over other methods. Compared to PromptIR \cite{potlapalli2023promptir} which adopts the same backbone (Restormer \cite{Zamir2021Restormer}), BaryIR obtains an average PSNR gain of 0.8 dB.  BaryIR also surpasses the recent InstructIR \cite{conde2024instructir} with an average PSNR gain of 0.42 dB and a 13.36 FID decline. Besides, BaryIR yields 1.11 dB and 0.97 dB gain on the dehazing and deraining tasks compared to InstructIR \cite{conde2024instructir}.

\textbf{Five degradations.} We further verify the effectiveness of BaryIR in a five-degradation scenario:  dehazing, deraining, denoising at level $\sigma=25$, deblurring, and low-light enhancement. As shown in Tab. \ref{air5}, BaryIR excels InstructIR \cite{conde2024instructir} with an average PSNR gain of 1.11 dB and a 15.53 FID reduction. Notably, BaryIR also proceeds InstructIR \cite{conde2024instructir} with 4.11 dB PSNR gain on the dehazing task, demonstrating its robustness to diverse degradations.

Fig. \ref{vair5} presents visual results under the five-degradation scenario. These examples show that, as compared to other methods, BaryIR not only consistently delivers balanced and superior performance in removing degradations (\textit{e.g.,} dense haze in the distant scene, severe real-world blur) but also produces results with better fine-grained structural contents (\textit{e.g.,} textures, colors). The underlying reason can be that BaryIR learns barycenters that encode common patterns of natural images, thereby effectively balancing multiple degradations and producing faithful results.
\subsection{Generalization to Real-world Scenarios}
\textbf{Single degradation.} We compare BaryIR with SOTA methods on unseen real-world haze O-HAZE \cite{ancuti2018haze}  and rain SPANet \cite{Wang_2019_CVPR} datasets using the five-degradation models. 
\begin{table}[!h]
	\caption{Generalization to unseen real-world O-HAZE \cite{ancuti2018haze}  and SPANet \cite{Wang_2019_CVPR} datasets with the five-degradation models.  The metrics are reported as PSNR($\uparrow$)/SSIM($\uparrow$)/LPIPS($\downarrow$)/FID($\downarrow$). }
	\label{unseen-real-tab}
	\vspace{-0.2cm}
	\resizebox{1.0\linewidth}{!}{
		\begin{tabular}{lcc}
			\toprule
			Method & \textit{Dehazing} on O-HAZE & \textit{Deraining} on SPANet\\
			\midrule
			Restormer \cite{Zamir2021Restormer}   &18.02/0.724/0.345/275.8&34.38/0.917/0.032/43.29\\
			IR-SDE \cite{luo2023image} &17.85/0.716/0.338/256.3&35.02/0.922/0.029/38.87\\
			\midrule
			PromptIR \cite{potlapalli2023promptir} & 18.38/0.730/0.336/260.1&35.34/0.938/0.026/33.12\\
			DA-CLIP \cite{luocontrolling}&18.22/0.725/0.323/242.5&35.65/0.942/0.026/26.96\\
			RCOT \cite{tang2024residualconditioned}& \underline{19.12}/\underline{0.745}/\underline{0.303}/\underline{216.8}&36.18/0.944/\underline{0.025}/\underline{25.58}\\
			InstructIR \cite{conde2024instructir}&18.85/0.738/0.308/236.5&\underline{36.42}/\underline{0.946}/0.028/30.54\\
			\midrule
			BaryIR &\textbf{22.93}/\textbf{0.792}/\textbf{0.256}/\textbf{173.2}&\textbf{39.15}/\textbf{0.971}/\textbf{0.014}/\textbf{16.85}\\
			\bottomrule
	\end{tabular}}
	\vspace{-0.3cm}
\end{table}
\begin{figure*}[!htbp]
	\setlength\tabcolsep{1pt}
	\renewcommand{\arraystretch}{0.75} % Default value: 1
	\centering
	\begin{tabular}{cccccc}
		\includegraphics[width=0.16\linewidth]{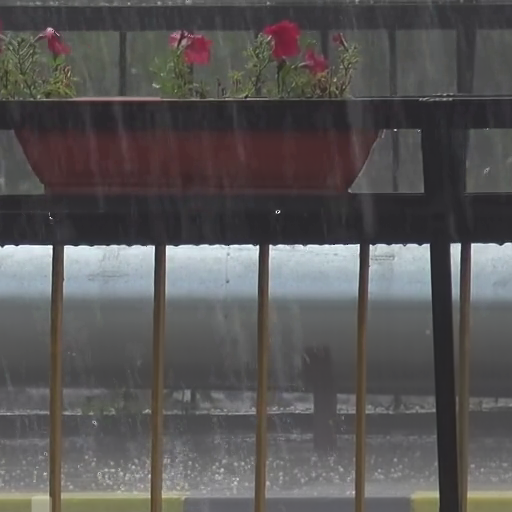}&
		\includegraphics[width=0.16\linewidth]{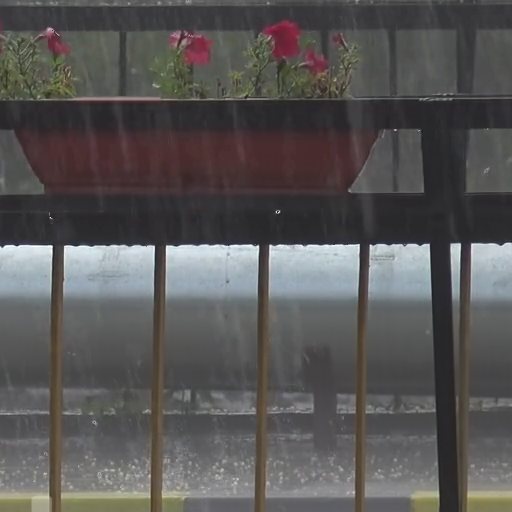}&
		\includegraphics[width=0.16\linewidth]{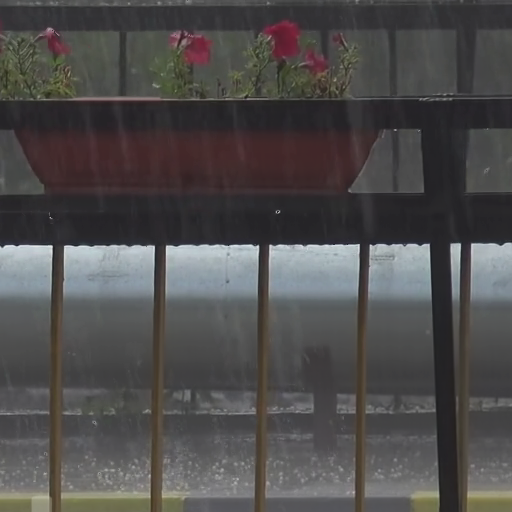}&
		\includegraphics[width=0.16\linewidth]{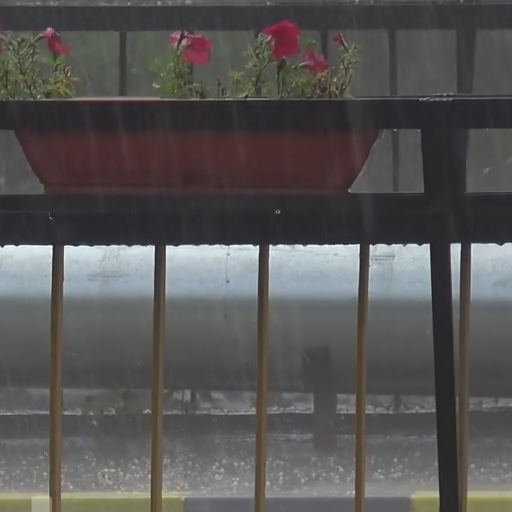}&
		\includegraphics[width=0.16\linewidth]{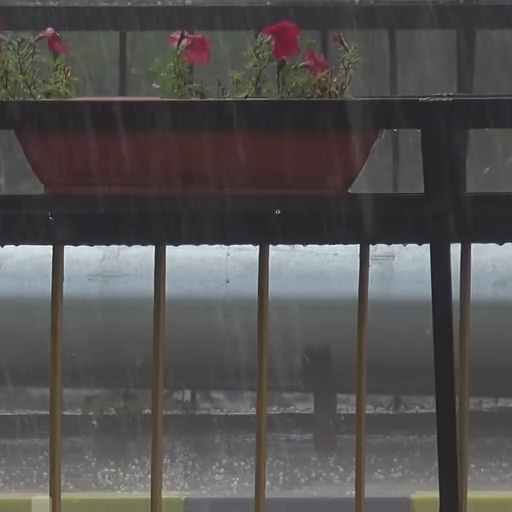}&
		\includegraphics[width=0.16\linewidth]{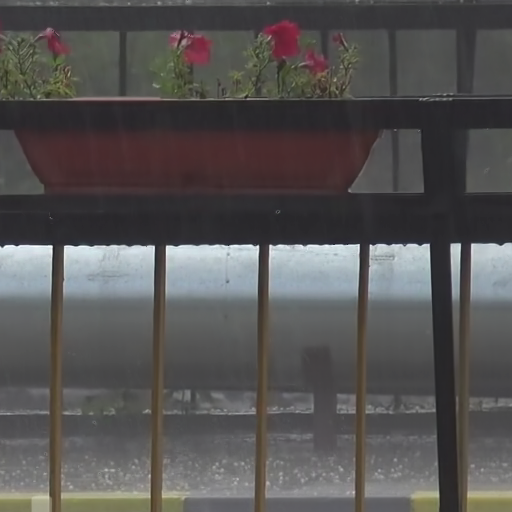}\\
		\includegraphics[width=0.16\linewidth]{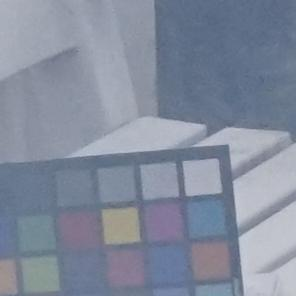}&
		\includegraphics[width=0.16\linewidth]{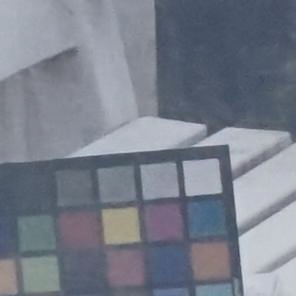}&
		\includegraphics[width=0.16\linewidth]{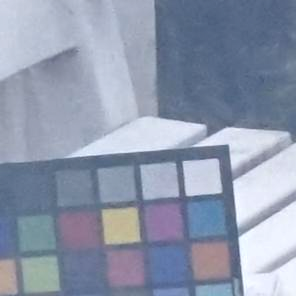}&
		\includegraphics[width=0.16\linewidth]{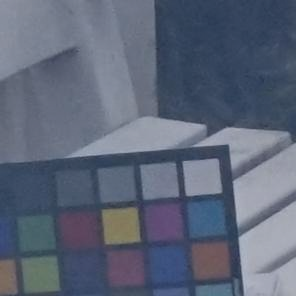}&
		\includegraphics[width=0.16\linewidth]{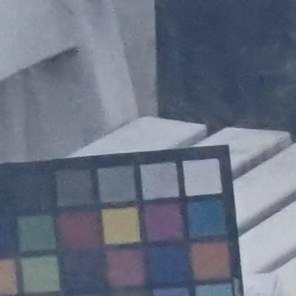}&
		\includegraphics[width=0.16\linewidth]{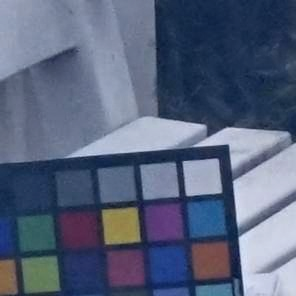}\\
		Degraded&PromptIR&DA-CLIP&RCOT&InstructIR&BaryIR\\
	\end{tabular}
	\caption{Visual examples of generalization evaluation with five-degradation models on unseen real-world O-HAZE \cite{ancuti2018haze}  and SPANet \cite{Wang_2019_CVPR}.}
	\vspace{-0.3cm}
	\label{unseen-real-fig}
\end{figure*}

Tab. \ref{unseen-real-tab} reports the quantitative results and shows that BaryIR yields PSNR gains of 3.81 dB on O-HAZE \cite{ancuti2018haze} and 2.73 dB on SPANet \cite{Wang_2019_CVPR} over the second-best methods. Fig. \ref{unseen-real-fig} displays the visual examples, in which the compared methods fail to remove the rain/haze or to restore image patterns properly. In contrast, BaryIR restores comparatively clear images with better visual contents, \textit{e.g.,} colors. These results reveal that BaryIR also delivers better generalization performance to unseen real-world data.
\begin{table}[!h]
	\caption{Generalization to real-world mix-degradation images from SPANet \cite{Wang_2019_CVPR} (haze and rain) and Lai \cite{lai2016comparative} (blur and noise). }
	\label{mixed-tab}
	\vspace{-0.2cm}
	\resizebox{1.0\linewidth}{!}{
		\begin{tabular}{lcccc}
			\toprule
			\multirow{2}{*}{Method} & \multicolumn{2}{c}{Haze and Rain}  &\multicolumn{2}{c}{Blur and Noise}\\
			&NIQE ($\downarrow$)&PIQE ($\downarrow$)&NIQE ($\downarrow$)&PIQE ($\downarrow$)\\
			\midrule
			Restormer \cite{Zamir2021Restormer}   &9.62&115.8&8.56&96.42\\
			IR-SDE \cite{luo2023image} &9.45&112.1&8.75&100.5\\
			\midrule
			PromptIR \cite{potlapalli2023promptir} &8.05&102.4&7.22&78.44\\
			DA-CLIP \cite{luocontrolling}&7.72&95.40&7.45&83.25\\
			RCOT \cite{tang2024residualconditioned}&\underline{7.20}&\underline{80.55}&6.46&70.24\\
			InstructIR \cite{conde2024instructir}&7.37&85.93&\underline{6.28}&\underline{62.18}\\
			\midrule
			BaryIR &\textbf{4.62}&\textbf{49.32}&\textbf{3.81}&\textbf{38.32}\\
			\bottomrule
	\end{tabular}}
	\vspace{-0.5cm}
\end{table}

\textbf{Mixed degradation.} Additionally, we evaluate on 49 mixed-degradation images collected from real-world datasets Lai \cite{lai2016comparative} (blur and noise),  SPANet \cite{Wang_2019_CVPR} (rain and haze), using no-reference metrics NIQE \cite{mittal2012making} and PIQE \cite{venkatanath2015blind}  for evaluation.
\begin{figure*}[!htbp]
	\setlength\tabcolsep{1pt}
	\renewcommand{\arraystretch}{0.75} % Default value: 1
	\centering
	\begin{tabular}{cccccc}
		\includegraphics[width=0.15\linewidth]{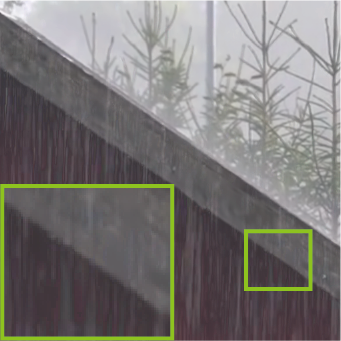}&
		\includegraphics[width=0.15\linewidth]{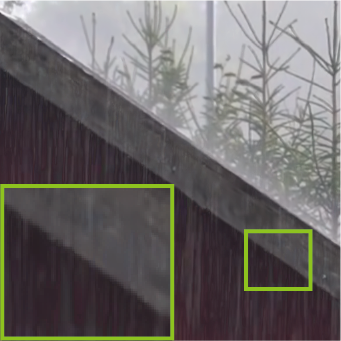}&
		\includegraphics[width=0.15\linewidth]{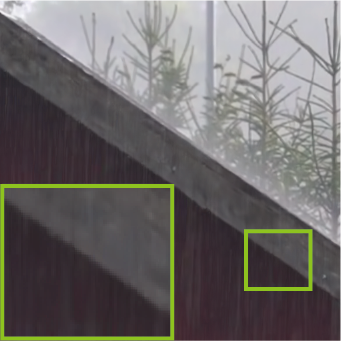}&
		\includegraphics[width=0.15\linewidth]{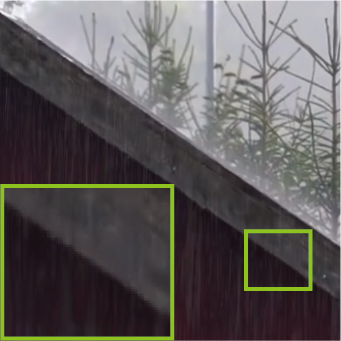}&
		\includegraphics[width=0.15\linewidth]{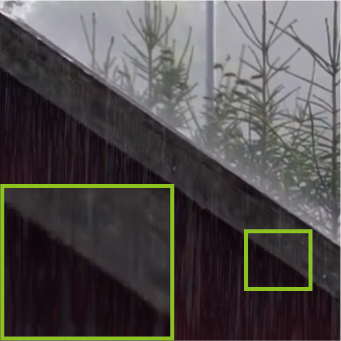}&
		\includegraphics[width=0.15\linewidth]{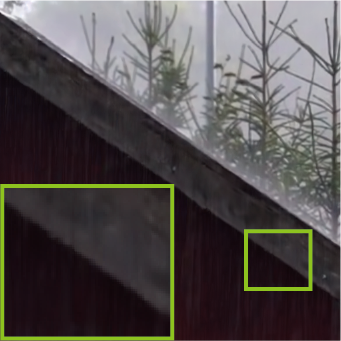}\\
		\includegraphics[width=0.15\linewidth]{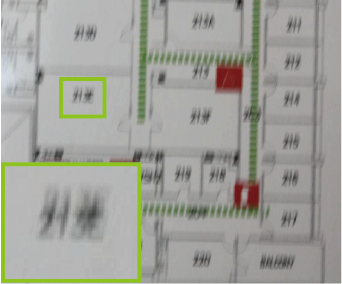}&
		\includegraphics[width=0.15\linewidth]{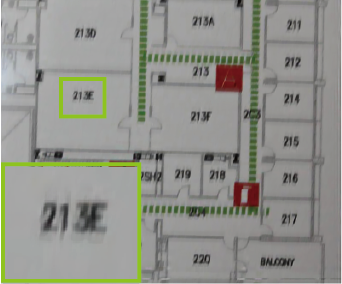}&
		\includegraphics[width=0.15\linewidth]{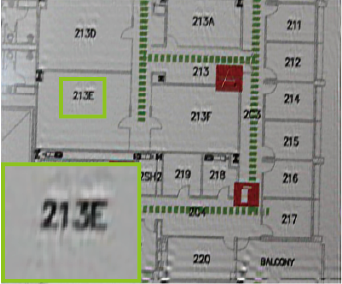}&
		\includegraphics[width=0.15\linewidth]{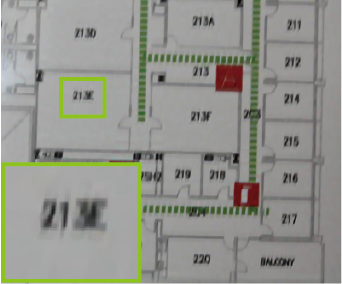}&
		\includegraphics[width=0.15\linewidth]{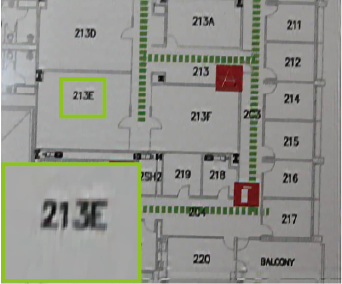}&
		\includegraphics[width=0.15\linewidth]{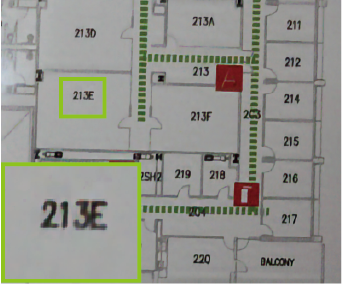}\\
		Degraded&PromptIR&DA-CLIP&RCOT&InstructIR&BaryIR\\
	\end{tabular}
	\vspace{-0.2cm}
	\caption{Visual examples on unseen real-world mixed-degradation images. Row 1: haze and rain. Row 2: blur and noise.}
	\vspace{-0.3cm}
	\label{mixed-fig}
\end{figure*}
Tab. \ref{mixed-tab} and Fig. \ref{mixed-fig} present the results, which show that BaryIR consistently outperforms other methods with significant quantitative and qualitative advantages in generalizing to real-world mixed-degradation images.

\subsection{Generalization to Unseen Degradation Levels}	
We evaluate the OOD performance on unseen degradation levels. Specifically, we train three-degradation models for dehazing (SOTS \cite{li2018benchmarking}), deraining (Rain100H \cite{yang2017deep}), and denoising with noise levels $\sigma\in\{15,25,50\}$ (BSD400 \cite{arbelaez2010contour} and WED \cite{ma2016waterloo}). We test the pre-trained models for deraining on Rain100L and denoising on BSD68 with unseen severe noise levels $\sigma=60$ and $\sigma=75$. 
\begin{table}[!h]
	\vspace{-0.2cm}
	\centering
	\caption{The OOD deraining results on Rain100L.}
	\label{unseen1}
	\vspace{-0.2cm}
	\resizebox{1.0\linewidth}{!}{
		\begin{tabular}{lcccc}
			\toprule
			Method & PSNR ($\uparrow$) & SSIM ($\uparrow$) & LPIPS ($\downarrow$)& FID ($\downarrow$)\\
			\midrule
			Restormer \cite{Zamir2021Restormer}   & 28.76&0.901&0.140&63.21\\
			IR-SDE \cite{luo2023image} &28.49&0.897&0.123&55.21\\
			\midrule
			PromptIR \cite{potlapalli2023promptir} & 31.82&0.931&0.078&38.41 \\
			DA-CLIP \cite{luocontrolling}&32.87&0.944&0.066&35.12\\
			RCOT \cite{tang2024residualconditioned}  &33.45&0.950&0.042&29.51\\
			InstructIR \cite{conde2024instructir} & \underline{33.89}&\underline{0.954}&\underline{0.033}&\underline{23.24}\\
			\midrule
			BaryIR &\textbf{36.69}&\textbf{0.975}&\textbf{0.018}&\textbf{10.28}\\
			\bottomrule
	\end{tabular}}
	\vspace{0.2cm}
	\centering
	\caption{The OOD denoising results on BSD68. The metrics are reported as PSNR($\uparrow$)/SSIM($\uparrow$)/LPIPS($\downarrow$)/FID($\downarrow$).}
	\label{unseen2}
	\vspace{-0.3cm}
	\resizebox{1.0\linewidth}{!}{
		\begin{tabular}{lcc}
			\toprule
			Method & $\sigma=60$ & $\sigma=75$\\
			\midrule
			Restormer \cite{Zamir2021Restormer}   &18.30/0.465/0.273/165.2&13.76/0.358/0.476/205.1\\
			IR-SDE \cite{luo2023image} &17.55/0.410/0.245/142.2&13.35/0.332/0.456/185.2                                              \\
			\midrule
			PromptIR \cite{potlapalli2023promptir} & 21.94/0.584/0.227/122.4&18.55/0.402/0.401/167.6\\
			DA-CLIP \cite{luocontrolling}&19.68/0.465/0.221/142.1&16.92/0.382/0.402/166.3\\
			RCOT \cite{tang2024residualconditioned}& 24.39/0.624/0.189/\underline{94.12}&19.32/0.454/0.388/160.3\\
			InstructIR \cite{conde2024instructir}& \underline{24.56}/\underline{0.626}/\underline{0.160}/98.46&\underline{19.55}/\underline{0.455}/\underline{0.374}/\underline{155.8}\\
			\midrule
			BaryIR &\textbf{26.83}/\textbf{0.749}/\textbf{0.134}/\textbf{74.63}&\textbf{22.85}/\textbf{0.507}/\textbf{0.324}/\textbf{116.6}\\
			\bottomrule
	\end{tabular}}
	\vspace{-0.3cm}
\end{table}
\begin{table*}[!htbp]
	\centering
	\caption{The results with different transport costs or representations for decoding. Metrics are reported as PSNR($\uparrow$)/LPIPS($\downarrow$). }
	\label{rep}
	\vspace{-0.2cm}
	\setlength{\tabcolsep}{12pt}
	\renewcommand{\arraystretch}{1}
	\resizebox{\textwidth}{!}{
		\begin{tabular}{l|cccccc|cc}
			\toprule
			Method& SOTS &Rain100L &  BSD68\textsubscript{$\sigma$=25} & GoPro& LOL &Average &O-HAZE &SPANet \\
			\midrule
			Original Rep.  									
			&24.09/0.065 &34.81/0.045&30.78/0.095&27.22/0.174&20.41/0.109&27.46/0.098&18.02/0.345&34.38/0.032\\
			Barycenter Rep.&30.27/0.015&37.23/0.025&31.05/0.088&28.05/0.155&22.86/0.096&29.89/0.076&22.04/0.278&38.53/0.022\\
			Original + Source-specific Rep. & 29.40/0.019 &36.23/0.027&30.88/0.093&27.40/0.170&21.78/0.105&29.14/0.083&19.84/0.295&36.22/0.029\\
			Barycenter + Source-specific Rep.  & \textbf{31.12}/\textbf{0.010}&\textbf{38.05}/\textbf{0.011}&\textbf{31.43}/\textbf{0.086}&\textbf{29.30}/\textbf{0.141}&\textbf{23.38}/\textbf{0.092}&\textbf{30.66}/\textbf{0.068}&\textbf{22.93}/\textbf{0.256}&\textbf{39.15}/\textbf{0.014}\\
			\midrule
			$c(z_k,z^B)$&28.45/0.022&36.44/0.033&30.62/0.093&27.51/0.165&22.15/0.108&29.03/0.084&21.48/0.287&36.89/0.027\\
			$c(z_k,z^B)+\mathcal L_{k}^{ctr}$&30.45/0.013&37.56/0.022&31.02/0.091&28.45/0.154&22.65/0.098&30.03/0.076&21.98/0.285&37.45/0.026\\
			$c(z_k,z^B)+\mathcal L_{k}^{ort}$&29.32/0.021&36.76/0.028&30.70/0.092&27.88/0.160&22.46/0.103&29.41/0.080&21.87/0.289&37.32/0.025\\
			$c(z_k,z^B)+\mathcal L_{k}^{ctr}+\mathcal L_{k}^{ort}$ & \textbf{31.12}/\textbf{0.010}&\textbf{38.05}/\textbf{0.011}&\textbf{31.43}/\textbf{0.086}&\textbf{29.30}/\textbf{0.141}&\textbf{23.38}/\textbf{0.092}&\textbf{30.66}/\textbf{0.068}&\textbf{22.93}/\textbf{0.256}&\textbf{39.15}/\textbf{0.014}\\
			\bottomrule
	\end{tabular}}
	\vspace{-0.5cm}
\end{table*}

From Tab. \ref{unseen1} and Tab. \ref{unseen2} we can see that BaryIR achieves superior quantitative advantages over other methods when generalizing to unseen degradation levels, \textit{e.g.,} 2.80 dB PSNR gain for deraining on Rain100L \cite{yang2017deep}, and 3.30 dB gain for denoising with severe unseen noise level $\sigma=75$ over InstructIR \cite{conde2024instructir}. These results reveal the generalizability of BaryIR in unseen real-world images and degradations, verifying the validity of using barycenters to encode multi-source unified representations for generalizable AIR.

\subsection{Ablation Studies and Model Analysis}
\noindent\textbf{Effect of the different latent representations.}
To investigate the effect of the representations in the barycenter and source-specific spaces, we compare models trained with different representations (denoted as Rep.) for decoding: 1) the original Rep.; 2) the barycenter Rep.; 3) aggregated original and the source-specific Rep.; 4) aggregated barycenter and source-specific Rep. (full model). Tab. \ref{rep} reports the results on five-degradation benchmark datasets and the generalization performance. We can observe that the barycenter Rep. alone can yield decent unified image restoration results and largely improve the model's generalizability. By aggregating Rep. from the barycenter and source-specific spaces, BaryIR achieves the best performance. The results verify the importance of both barycenter and source-specific Rep. for generalizable AIR.

\quad\\
\noindent\textbf{Effect of the transport cost terms in MLOT objective.}
We investigate the effect of transport cost terms in the MLOT objective, including the source-level contrastiveness term $\mathcal L_{k}^{ctr}$ and the barycenter-anchored orthogonality term $\mathcal L_{k}^{ort}$. We can observe from Tab. \ref{rep} that both terms bring non-trivial improvement to the performance. The best performance is achieved with two terms $\mathcal L_{k}^{ctr}$ and $\mathcal L_{k}^{ort}$ working together, particularly in terms of the generalization results.

\begin{figure}[!t]
	\centering
	\includegraphics[width=1\linewidth]{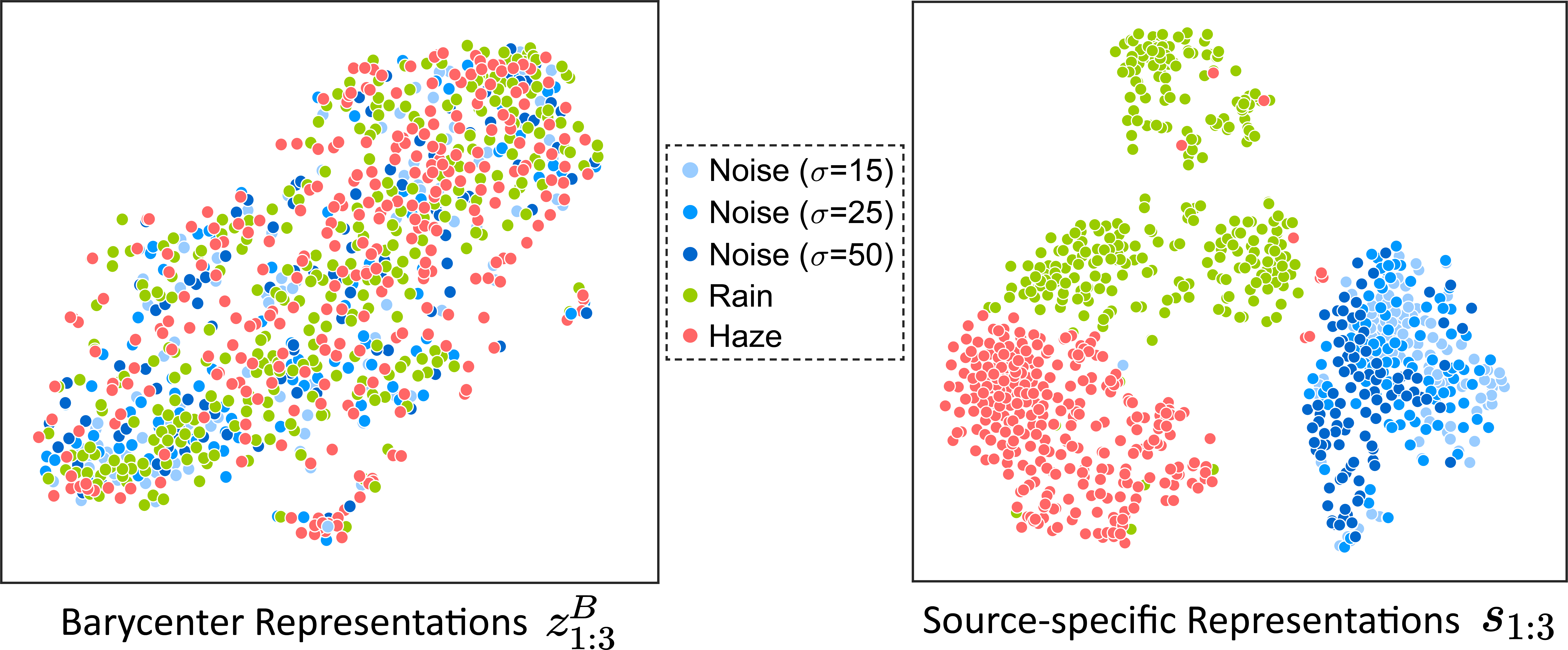}
	\caption{The t-SNE visualization of different representations.}
	\label{vis_rep}
	\vspace{-0.6cm}
\end{figure}

\quad\\
\noindent\textbf{The t-SNE visualization of the barycenter and source-specific representations.} Given the motivation of using barycenter for degradation-agnostic features and source-specific representations for degradation-specific semantics, we present a t-SNE plot across degradations. 300 noisy images (100 each for $\sigma=15, \sigma=25, \sigma=50$), 300 rainy images, and 300 hazy images are used.  As shown in Fig. \ref{vis_rep}, the barycenter representations capture degradation-agnostic features, and the source-specific ones are separated according to the degradation types, aligning with our motivation. 
\vspace{-0.3cm}
\section{Conclusion}
\vspace{-0.1cm}
This paper proposed a BaryIR framework for the AIR problem. With the dual reformulation of the multi-source latent OT barycenter problem, we learned an NN-based barycenter map to transport the representations to the barycenter space for unified encoding and exploited the source-specific subspaces for degradation-specific semantics.  By aggregating representations from both spaces, BaryIR can produce generalizable AIR solutions. Extensive experiments demonstrated the effectiveness of BaryIR for unified image restoration, especially in terms of its generalizability in real-world and unseen degradations. In the future, we aim to establish barycenter-driven unified representation for multi-modal signals, \textit{e.g.,} text, image, and audio, which may depend on the design of transport costs.

{
	\small
	\bibliographystyle{ieeenat_fullname}
	\bibliography{main}
}
%\input{Supp}
\end{document}